\documentclass[11pt]{article}
\PassOptionsToPackage{numbers,sort&compress}{natbib}
\usepackage[margin=1in]{geometry}
\usepackage{microtype}
\usepackage{amsmath,amssymb}
\usepackage{graphicx}
\usepackage{booktabs}
\usepackage{natbib}
\usepackage{hyperref}
\usepackage{xcolor}
\usepackage[utf8]{inputenc} % allow utf-8 input
\usepackage[T1]{fontenc}    % use 8-bit T1 fonts
\usepackage{hyperref}       % hyperlinks
\usepackage{url}            % simple URL typesetting
\usepackage{booktabs}       % professional-quality tables
\usepackage{amsfonts}       % blackboard math symbols
\usepackage{nicefrac}       % compact symbols for 1/2, etc.
\usepackage{microtype}      % microtypography
\usepackage{xcolor}         % colors
\usepackage{amsmath}
\usepackage{booktabs}
\usepackage{mathtools}
\usepackage{algorithm}
\usepackage{algpseudocode}
\usepackage{enumitem}
\usepackage{subcaption}
\usepackage[capitalize,nameinlink]{cleveref}

\title{On the Nonlinearity of Learning Rate Scaling for LLM Training}

\author{%
    Zaiwen Yang \\
    Tsinghua University \\
    \and
    Huaqing Zhang \\
    Tsinghua University \\
    \and
    Jing Xu \\
    Tsinghua University \\
    \and
    Jingzhao Zhang\thanks{Correspondance to yangzw23@mails.tsinghua.edu.cn and jingzhaoz@tsinghua.edu.cn} \\
    Tsinghua University, Shanghai Qi Zhi Institute \\
}
\date{}

\newif\ifcomments
\commentstrue

\definecolor{huaqing}{RGB}{0,110,80}

\begin{document}

\maketitle

\begin{abstract}
Learning-rate transfer can reduce the cost of training large language models: instead of sweeping learning rates at target scale, practitioners extrapolate from smaller runs. Existing approaches often assume that the optimal learning rate follows a log-linear scaling law in data scale and model size. We carefully examine and evaluate this scaling law. In our empirical study of GPT-2--style models from 22M to 707M parameters trained on 5B to 100B tokens, the optimal learning rate develops upward curvature at larger scales, leading to inaccurate extrapolation. We find that this curvature largely disappears when learning rates are replaced by effective learning rate (the step size in normalized weight space), and when data $D$ extrapolation is used instead of model size $N$ extrapolation. Next, we explain nonlinearity in scaling: weight-norm converges to equilibrium slower when optimal learning is small, requiring a larger step size to reduce the transient phase. Experiments with AdamH, which directly controls the effective learning rate, further support this explanation. 
\end{abstract}

\section{Introduction}

Efficiently training large language models requires carefully tuned hyperparameters, yet sweeping these hyperparameters directly at the target scale is often prohibitively expensive. \emph{Hyperparameter transfer}---predicting effective hyperparameters at large scale from inexpensive smaller-scale runs---has therefore become a central tool in modern model development \citep{mup, li2025steplaw}.

The learning rate is a particularly consequential hyperparameter, as it directly affects both training efficiency and stability. Accordingly, many prior works have studied the transfer of learning rates across training settings. 
% One is $\mu$P style where the optimal learning rate is controlled~\citep{mup}. From theoretical scaling arguments, these works derive width scaling hyperparameter transfer rules, which is later extended to other settings, including depth scaling, alternative optimizers, and low-precision training~\citep{ bordelon2024depthwise, blakeu2025umup, everett2024scaling}. We see that these approaches primarily address transfer along the model-size axis $N$.
One line of work is $\mu$P~\citep{mup}: under this parameterization, the optimal learning rate remains stable across width scaling. Subsequent works have extended this framework to other settings such as depth scaling \citep{bordelon2024depthwise, blakeu2025umup, everett2024scaling}. These approaches mostly focus on learning-rate transfer along the model-size scaling axis.

Some recent work shows that $\mu$P style analysis, though very elegant, can be inaccurate when weight norm changes during training~\citep{kosson2025wdimportant, zhou2026setlr}. 
% More recent studies instead model the scaling behavior of the optimal learning rate $\eta^\ast$ as a function of data scale $D$ and model size $N$. 
As a complement to $\mu$P-style analyses, and to enable transfer along the data-size axis, another line of work directly models the optimal learning rate  $\eta^\ast$ as a function of the data scale $D$ and model size $N$.
The most common assmuption is that $\eta^\ast$ follows a log-linear, or power-law, relationship in $D$ and $N$~\citep{kaplan2020openaiscaling,bjorck2025microsoftlaw,li2025steplaw}:
\begin{equation*}
    \log \eta^\ast(D, N) = a \log D + b \log N + c.
\end{equation*}

Existing studies often fit the scaling law under a fixed search protocol and assess whether it extrapolates to larger model or data scales. We examine this paradigm from two perspectives. First, obtaining an accurate scaling-law fit can itself require substantial compute, but this cost is often omitted when comparing transfer strategies. To enable a fair comparison, we propose measuring distinct power-law relationships under the same compute budget. Second, rather than jointly fitting the dependence on the data scale $D$, the model scale $N$, and the learning rate, we further quantify the extrapolation efficiency of different transfer strategies on two choices of the transfer axis, the data scale $D$ versus the model scale $N$, and two parameterizations of the learning rate, the learning rate versus the effective learning rate. Here the effective learning rate is defined as~\citep{wen2025hyperball}
\[
\eta_\mathrm{eff}(t)=\|\hat{w}_{t+1}-\hat{w}_t\|_2,
\]
where $\hat{w}=w/\|w\|_2$ denotes the normalized weight direction.

Motivated by the above two perspectives, we conduct a systematic empirical study of learning-rate transfer across model and data scales.   
Specifically, we train GPT-2--style models \citep{radford2019gpt2} ranging from 22M to 707M parameters on 5B to 100B tokens, and hold out the largest 707M model as the test set for evaluating optimal learning-rate prediction.
We release a comprehensive dataset \footnote{\url{https://huggingface.co/datasets/neurips2026lrtransfer/LR-Transfer-Trajectory}} spanning model sizes, data scales, and learning rates, with dense coverage of loss trajectories and optimization statistics. 
 
Our main findings from these experiments are as follows.
\begin{itemize}[itemsep=0pt,leftmargin=*,topsep=1pt]

    \item We find that the optimal learning rate $\eta^\ast$ at large data scale $D$ and model size $N$ is higher than predicted by log-linear extrapolation from smaller $D$ and $N$. In contrast, the effective learning rate $\eta_{\mathrm{eff}}$ exhibits substantially more linear scaling behavior  (\cref{sec:main_results}).

    \item 
    %Under controlled compute budgets, we quantitatively compare extrapolation along the data-size axis with extrapolation along the model-size axis. We find that extrapolating over data size can be much more compute-efficient, yielding near optimal training compute with less than $20\%$ compute spent for scaling law-fitting (details in Table \ref{tab:prediction_table}).
    % We further compare scaling-law fits based on learning rates $\eta$ against those based on effective learning rates $\eta_{\mathrm{eff}}$. finding that extrapolating along $D$ with $\eta_{\mathrm{eff}}$ yields accurate predictions even at the low end of the compute-budget range. 
    Under controlled compute budgets, we quantitatively compare the extrapolation performance along the data-size and model-size axes.
    We find that fitting learning-rate scaling law along the data-size axis can yield much more accurate lr predictions: it incurs only $2\%$ additional training compute relative to using the true optimal learning rate, and requires less than $20\%$ of the compute of hyperparameter sweeping at the target scale (details in Table~\ref{tab:prediction_table}).

    \item By analyzing the implicit dynamics of AdamW and the mapping between $\eta$ and $\eta_{\mathrm{eff}}$, we explain the upward curvature observed in $\eta^\ast$. We further verify the linear scaling behavior of $\eta_{\mathrm{eff}}$ by explicitly controlling it during training with the AdamH optimizer \citep{wen2025hyperball} (\cref{sec:effective_lr_schedule}).
\end{itemize}

\section{Related Works}

% All hyperparameters settings

% \begin{itemize}
%     \item Background: Hyperparameter Transfer and Scaling
%     \item Scaling Laws for Learning Rates: Microsoft, Step Law, AI Lab. Lack of existing works: focus primarily on fit quality.
%     \item Effective Learning Rate: hyperball descent, Kosson. 
% \end{itemize}

\paragraph{Transferring learning rate across training scales.}

Hyperparameter tuning is crucial for the efficient training of large language models. In this work, we focus on the transferring the learning rate. One common approach is to fit parametric scaling laws, which model the optimal learning rate as a power-law function of the training scale, such as model size \citep{kaplan2020openaiscaling, porian2024porianlaw}, training loss \citep{wang2024meituanlaw}, or compute budget \citep{deepseekai2024deepseekscaling}. 
Later work refines this formulation by making the optimal learning rate depend on both model size and data size \citep{bjorck2025microsoftlaw, zhou2026setlr}, and \citet{li2025steplaw} further study the joint tuning of learning rate and batch size.
However, the extrapolation performance of these power-law relations, as well as the range of training scales over which they hold, remains insufficiently evaluated. We aim to address this gap.

% Kaplan, Deepseek: C. Microsoft, AILab: N, D; Steplaw, joint tuning of lr and bs. However, the range that the power law relation holds and whether could be better way is under explored. 

Another line of work for transferring learning rates across model scales is the theoretically motivated Maximal Update Parameterization ($\mu$P) \citep{mup}. 
Later works extend the original width scaling to other training settings, such as depth scaling, different optimizers, and low-precision training \citep{dey2025completep, bordelon2024depthwise, Greg2021TPIV, blakeu2025umup, everett2024scaling}.
However, $\mu$P focuses on model scaling and does not directly cover transfer to larger data scales.

\paragraph{Effective learning rate.}

Many weight matrices in mordern architectures are scale-invariant due to normalization layers~\citep{ioffe2015batch, ba2016layer}: rescaling such weights by a positive scalar leaves the network output, and hence the loss, unchanged~\citep{van2017l2, hoffer2018norm}. As a result, the relevant training dynamics is governed primarily by changes in weight direction rather than weight scale. Previous work shows that the effective learning rate is jointly determined by the learning rate and weight decay \citep{li2020reconciling, kosson2024rotational}.
This perspective has further motivated new optimizer designs that explicitly control the weight norm and update norm \citep{loshchilovngpt, funemotron, xie2026controlled}.
In this work, we further show that the effective learning rate, rather than the learning rate, is the right quantity to consider for hyperparameter transfer.

We discuss more related works on hyperparameter tuning and transfer in \cref{app:additional_related_work}.

\section{Power Laws for Optimal Learning Rates}
\label{sec:preliminaries}

% We train language models with AdamW~\citep{loshchilov2018decoupled}. Our experiments focus on the learning rate $\eta$ as the hyperparameter of interest. All other AdamW hyperparameters, including $\beta_1$, $\beta_2$, $\epsilon$, and $\lambda$, are fixed. 

Our experiments focus on the learning rate $\eta$ as the hyperparameter of interest.  Let $D$ denote the number of training tokens and $N$ denote the number of model parameters. Throughout this work, we fix the batch size $B$, so that the total number of training tokens satisfies $D=BT$, where $T$ is the total number of training steps. We define $\eta^\ast(D,N)$ as the optimal learning rate that minimizes the target loss $\mathcal{L}$ after training on $D$ tokens with a model of scale $N$:
\begin{equation}
    \eta^\ast(D,N) \coloneq \mathop{\arg\min}_{\eta} \mathcal{L}\left(w_T(\eta;D,N)\right).
    % \quad D=BT.
\end{equation}
Previous works assume that the optimal learning rate follows a log-linear scaling law with respect to scale  \citep{bjorck2025microsoftlaw, li2025steplaw, zhou2026setlr}:
\begin{equation}
    \log \eta^\ast(D,N) = a \log D + b \log N + c,
\end{equation}
where $a,b,c$ are constants determined empirically. This formulation implies that the learning rate obeys a power law relationship $\eta^\ast(D,N) \propto D^aN^b$,
% which is widely used for hyperparameter transfer across scales \citep{bjorck2025microsoftlaw, li2025steplaw, zhou2026setlr}.

\paragraph{Effective Learning Rate}
In this work, we also study the scaling behavior of the effective learning rate.
Modern neural network architectures such as Transformers \citep{NIPS2017attentionallyouneed} are often scale-invariant with respect to their weights. In particular, for most matrix weights $w$, due to normalization layers, the loss remains unchanged under rescaling~\citep{van2017l2, hoffer2018norm}:
\begin{equation}
    \mathcal{L}(w) = \mathcal{L}(\alpha w), \quad \forall \alpha > 0.
\end{equation}
As a result, the optimization dynamics depend primarily on the direction of the weights rather than their magnitude. From this perspective, we define the effective learning rate as the step size in the normalized parameter space~\cite{wen2025hyperball}:
\begin{equation}
    \label{eq:efflr_definition}
    \eta_{\mathrm{eff}}(t)=\left\|\hat{w}_{t+1} - \hat{w}_t\right\|_2,
\end{equation}
where $\hat{w}_t = w_t / \|w_t\|_2$ denotes the normalized weight vector.

Our experiments show that, when the learning rate is fixed, the effective learning rate varies slowly over the course of training; see \autoref{fig:efflr_weight_norm_vs_t}. This suggests that $\eta_{\mathrm{eff}}(t)$ can be approximated by a constant during training. For a given dataset size $D$, model size $N$, and learning rate $\eta$, we therefore summarize the effective learning rate by its average over training:
\begin{equation}
    \label{eq:efflr_avg_over_t}
    \overline{\eta}_{\mathrm{eff}}(\eta;D,N)
    \coloneqq
    \frac{1}{T}
    \sum_{t=0}^{T-1}
    \eta_{\mathrm{eff}}(t;\eta,D,N),
    % \quad D=BT,
\end{equation}
where $T$ is the total number of training steps.
Hence, we define the optimal effective learning rate at scale $(D,N)$ as the average effective learning rate induced by the optimal learning rate $\eta^\ast(D,N)$:
\begin{equation}
    \eta_{\mathrm{eff}}^\ast(D,N)
    \coloneqq       
    \mathop{\arg\min}_{\overline{\eta}_{\mathrm{eff}}}
    \mathcal{L}\left(\hat{w}_T(\overline{\eta}_{\mathrm{eff}};D,N)\right).
    % \quad D=BT.
\end{equation}

\section{Experiment Design}
\label{sec:experiment_design}
In this work, we evaluate four log-linear scaling laws for the optimal learning rate and the optimal effective learning rate. 
First, for each fixed model size $N$, we study how the optimal learning rate scales with data size:
\begin{equation}
\label{eq:eta_d_scaling_law}
\log \eta^\ast(D, N) = a_N \log D + b_N.
\end{equation}
Further, for each fixed data size $D$, we study how the optimal learning rate scales with model size:
\begin{equation}
\label{eq:eta_n_scaling_law}
\log \eta^\ast(D, N) = c_D \log N + d_D.
\end{equation}
Similarly, we evaluate the corresponding scaling laws for the optimal effective learning rate. For each fixed model size $N, D$, we consider
\begin{equation}
\label{eq:eff_eta_d_scaling_law}
\log \eta^\ast_{\mathrm{eff}}(D, N) = \tilde{a}_N \log D + \tilde{b}_N,
\end{equation}
% and for each fixed data size $D$, we consider
\begin{equation}
\label{eq:eff_eta_n_scaling_law}
\log \eta^\ast_{\mathrm{eff}}(D, N) = \tilde{c}_D \log N + \tilde{d}_D.
\end{equation}

We estimate these quantities from training runs over a grid of model sizes, data budgets, and learning rates.
For each $(D,N)$, we estimate $\eta^\ast(D,N)$ by fitting a cubic polynomial to validation loss as a function of $\log \eta$ (\autoref{fig:val_loss_vs_log2lr}). Similarly, we estimate $\eta_{\mathrm{eff}}^\ast(D,N)$ by fitting a cubic polynomial to validation loss as a function of $\log \eta_{\mathrm{eff}}$ (\autoref{fig:val_loss_vs_log2efflr}).

\begin{figure}[ht]
    \centering
    \includegraphics[width=0.60\linewidth]{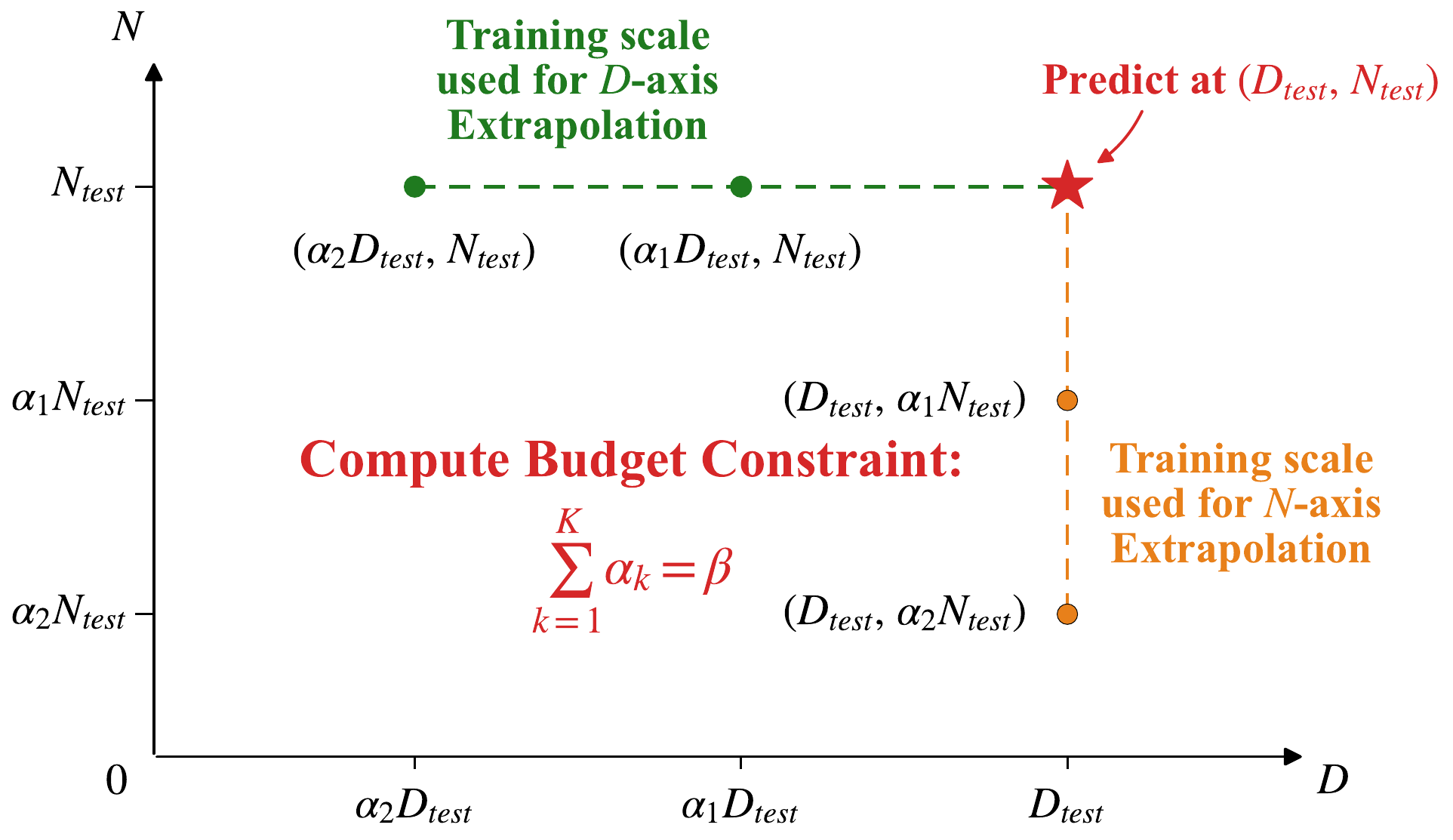}
    \caption{Compute-constrained extrapolation: $D$-axis extrapolation uses reduced data with the full model, while $N$-axis extrapolation uses smaller models with full data, both under a fixed compute budget.}
    \label{fig:how_to_fit}
\end{figure}

\paragraph{Experimental Setup.}
We train GPT-2 \citep{radford2019gpt2} models ranging from 22M to 707M parameters on FineWeb-100B~\citep{penedo2024fineweb} with data budgets from 5B to 100B tokens in 2.5B-token increments, using a warmup--steady--decay (WSD) schedule~\citep{hu2024minicpm}. 
We train language models with AdamW~\citep{loshchilov2018decoupled} and a fixed batch size of 0.52M tokens. All other AdamW hyperparameters except the learning rate, including $\beta_1$, $\beta_2$, $\epsilon$, and weight decay $\lambda$, are fixed. 
For each model size and data budget, we sweep the learning rate over a logarithmic grid with $\log_2 \eta \in [-16,-7]$ and unit increments. Further experimental details are provided in \autoref{app:experimental_setup}.

For each configuration $(D,N,\eta)$, we record the validation loss trajectories and key optimization statistics, including effective learning rates, weight norms $\|w_t\|_2$, gradient norms $\|g_t\|_2$, update norms $\|w_{t+1}-w_t\|_2$, and the norm of the AdamW update $\|u_t\|_2$. These trajectories and statistics are then used to estimate $\eta^\ast(D,N)$ and $\eta^\ast_{\text{eff}}(D,N)$ as described in \cref{sec:preliminaries}. The $\eta_{\text{eff}}$ is averaged on all weights except the embedding layer before averaging over $t$ in~\autoref{eq:efflr_avg_over_t}.

Our dense data-budget and learning-rate sweeps enable interpolation along both the $D$- and $\eta$-axes, densifying the $(D,\eta^\ast)$ for each model size $N$ without additional training runs.

\begin{figure}[ht]
    \centering
    \includegraphics[width=1.0\linewidth]{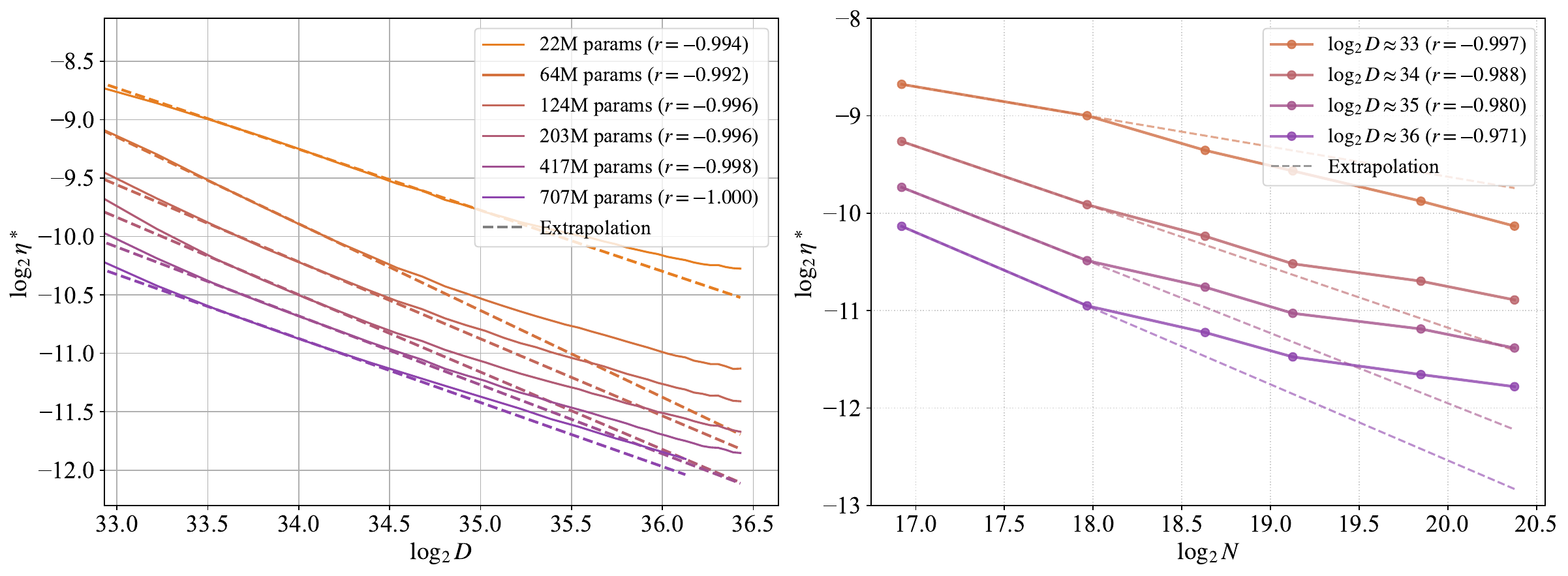}
    \caption{Log-log relationships for the optimal \textbf{learning rate $\eta^{\ast}$} as a function of training steps $D$ and model size $N$. While approximately linear over limited ranges, systematic curvature appears at larger scales, indicating that log-linearity is only a local property. The extrapolation line is fitted by $D \in [20\mathrm{B},40\mathrm{B}]$ or $N \in \{22\mathrm{M},64\mathrm{M}\}$, to evaluate extrapolation performance from small-scale fitting. $r$ refers to the Pearson correlation coefficient.}
    \label{fig:eta_combined}
\end{figure}

\begin{figure}[ht]
    \centering
    \includegraphics[width=1.0\linewidth]{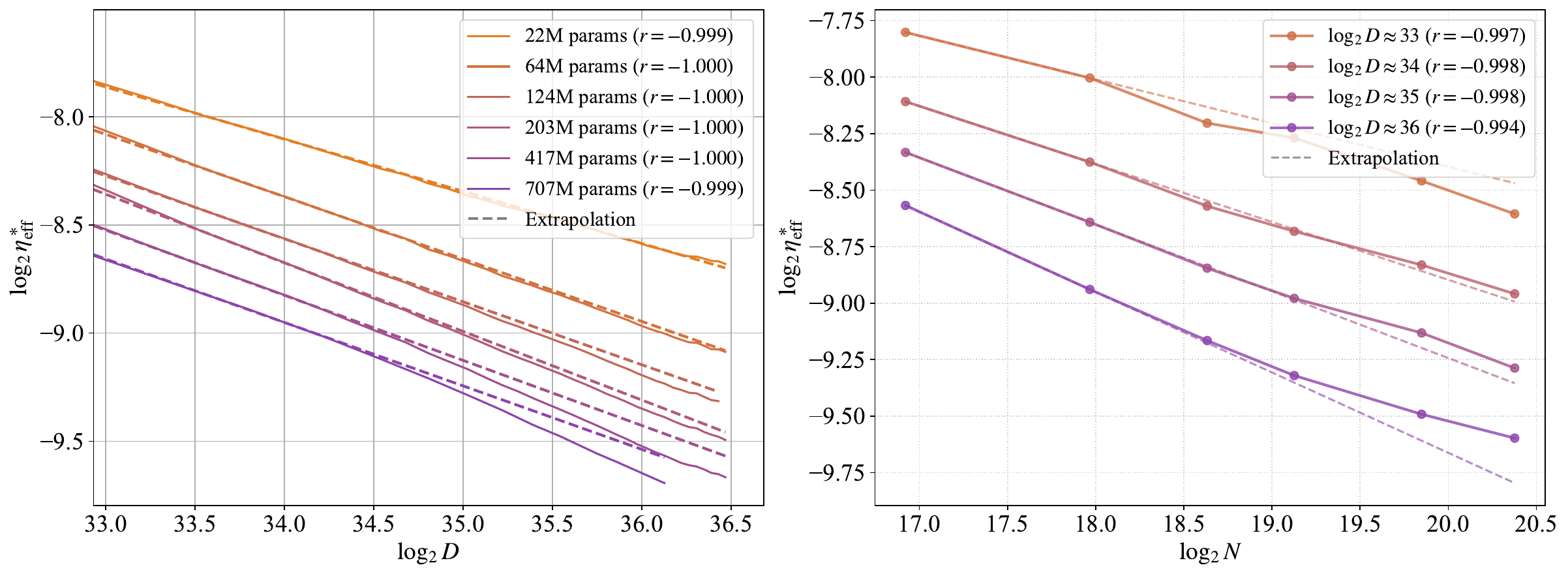}
    \caption{Log-log relationships for the optimal \textbf{effective learning rate $\eta_{\mathrm{eff}}^{\ast}$} as a function of $D$ and $N$. Compared with $\eta^{\ast}$, the trends remain closer to linear across scales, indicating more stable scaling behavior. The extrapolation line is fitted by $D \in [20\mathrm{B},40\mathrm{B}]$ or $N \in \{22\mathrm{M},64\mathrm{M}\}$, to evaluate extrapolation performance from small-scale fitting. $r$ refers to the Pearson correlation coefficient.}
    \label{fig:eff_eta_combined}
\end{figure}

\paragraph{Evaluating Extrapolation Performance.}

We evaluate each scaling law on a test set $\mathcal{S}_{\mathrm{test}} \subseteq \{(D,N,\eta^\ast(D,N))\}$. For each target tuple $(D,N,\eta^\ast(D,N)) \in \mathcal{S}_{\mathrm{test}}$, we define the reference compute $C(D,N) = 6ND$, corresponding to one full training run of a dense Transformer \citep{NIPS2017attentionallyouneed} with $N$ parameters on $D$ tokens~\citep{kaplan2020openaiscaling}. Given a compute-budget ratio $\beta > 0$, we construct a distinct training set $\mathcal{S}_{\mathrm{train}}(D,N)$ using at most $\beta C(D,N)$ total compute, fit each candidate scaling law on it, and evaluate the resulting prediction $\hat{\eta}^\ast_{\beta}(D,N)$ against the held-out empirical optimum $\eta^\ast(D,N)$.

We first define the \textit{out-of-distribution coefficient of determination} as
\begin{equation}
R^2_{\text{OOD}}(\beta) = 
1 - 
\frac{\sum_{(D,N)\in \mathcal{S}_{\text{test}}} 
\left( \log \eta^\ast(D,N) - \log \hat{\eta}^\ast_{\beta}(D,N) \right)^2}
{\sum_{(D,N)\in \mathcal{S}_{\text{test}}} 
\left( \log \eta^\ast(D,N) - \overline{\log \eta^\ast} \right)^2},
\end{equation}
where $\overline{\log \eta^\ast} = \frac{1}{|\mathcal{S}_{\text{test}}|} \sum_{(D,N)\in \mathcal{S}_{\text{test}}} \log \eta^\ast(D,N)$, measuring predictive accuracy in log-space on unseen scales. We remark that $R^2_{\text{OOD}}$ can be negative, since the extrapolated prediction $\log \hat{\eta}^\ast_{\beta}(D,N)$ may deviate substantially from the true optimum.

While $R^2_{\text{OOD}}$ captures statistical accuracy, the practical value of a scaling law lies in reducing the compute needed to identify near-optimal hyperparameters. We thus define the \textit{extra compute ratio} (ECR) to quantify the excess computation incurred by using predicted rather than optimal learning rates:
\begin{equation}
\mathrm{ECR}(\beta) = 
\frac{\sum_{(D,N)\in \mathcal{S}_{\text{test}}} \Delta C(D,N)}
{\sum_{(D,N)\in \mathcal{S}_{\text{test}}} C(D,N)},
\end{equation}
where we approximately set $C=6ND$, and $\Delta C(D,N)$ is the additional compute required for $\hat{\eta}^\ast_{\beta}(D,N)$ to match the validation loss attained by $\eta^\ast(D,N)$. Together, $R^2_{\text{OOD}}$ and $\mathrm{ECR}$ provide complementary views of extrapolation performance.

\paragraph{Test Set and Compute-constrained Training Sets.}
In our experiments, we fix $N_{\text{test}}$ to our largest model (hidden dimension 2048, 707M parameters) and vary $D_{\text{test}} \in [50\text{B}, 100\text{B}]$ tokens. We construct $\mathcal{S}_{\text{train}}$ via two strategies (\autoref{fig:how_to_fit}), both parameterized by coefficients $\{\alpha_k\}_{k=1}^K$:

\begin{itemize}[leftmargin=*]
    \item \textbf{Data-axis ($D$) extrapolation:} $\mathcal{S}_{\text{train}}^{(D)} = \{(\alpha_k D_{\text{test}}, N_{\text{test}})\}_{k=1}^K$, testing scaling laws \autoref{eq:eta_d_scaling_law} and \autoref{eq:eff_eta_d_scaling_law}.
    \item \textbf{Model-axis ($N$) extrapolation:} $\mathcal{S}_{\text{train}}^{(N)} = \{(D_{\text{test}}, \alpha_k N_{\text{test}})\}_{k=1}^K$, testing scaling laws \autoref{eq:eta_n_scaling_law} and \autoref{eq:eff_eta_n_scaling_law}.
\end{itemize}

Then the total training compute is $6 N_{\text{test}} D_{\text{test}} \sum_k \alpha_k$ for both strategies. Enforcing the budget $\beta C(D_{\text{test}}, N_{\text{test}})$ yields the constraint $\sum_{k=1}^K \alpha_k = \beta$, after which the fitted scaling law is extrapolated to predict $\eta^\ast(D_{\text{test}}, N_{\text{test}})$.

\paragraph{Estimating the Extra Compute Ratio.}

To compute $\Delta C(D_{\text{test}}, N_{\text{test}})$, we model the loss trajectory under the predicted learning rate $\hat{\eta}^\ast_{\beta}$ as a power law~\citep{kaplan2020openaiscaling} $L_{\hat{\eta}}(D) = L_0 + A D^{-\gamma}$, fitted from the validation-loss trajectory and anchored at $L_{\hat{\eta}}(D_{\text{test}}) = L_{\text{real}}(D_{\text{test}}, N_{\text{test}}, \hat{\eta}^\ast_{\beta})$. Letting $L^\ast(D_{\text{test}}, N_{\text{test}})$ denote the validation loss under $\eta^\ast$, we solve
$L_{\hat{\eta}}(D_{\text{test}} + D_{\text{extra}}) = L^\ast(D_{\text{test}}, N_{\text{test}})$
for $D_{\text{extra}}$, the additional tokens needed to recover the optimal loss. The extra compute is then $\Delta C(D_{\text{test}}, N_{\text{test}}) = 6 N_{\text{test}} D_{\text{extra}}$.

\section{Main Results}
\label{sec:main_results}
\paragraph{Deviation from Log-linear Scaling.}
We begin by examining whether the optimal learning rate follows the commonly assumed log-linear scaling laws. 

\autoref{fig:eta_combined} shows the relationship between the optimal learning rate $\eta^{\ast}$ and both training steps $D$ and model size $N$ in log-log space. 
While the trends appear approximately linear over moderate ranges, systematic upward curvature emerges as $D$ and $N$ increase. 
This deviation indicates that the log-linear scaling law is only locally valid and does not extend globally.

In contrast, \autoref{fig:eff_eta_combined} presents the corresponding results for the effective learning rate $\eta_{\mathrm{eff}}^{\ast}$. 
Compared to $\eta^{\ast}$, the effective learning rate shows improved linearity with higher Pearson correlation, particularly at larger scales. 
This suggests that $\eta_{\mathrm{eff}}^{\ast}$ provides a more stable parameterization on scaling.

\paragraph{Quantifying Extrapolation Error  Across Different Strategies.}

We next quantify how deviations from log-linearity affect extrapolation performance using the out-of-distribution coefficient of determination $R^2_{\text{OOD}}$.

\begin{figure}[htbp]
    \centering
    \begin{subfigure}{0.48\textwidth}
        \centering
        \includegraphics[width=\linewidth]{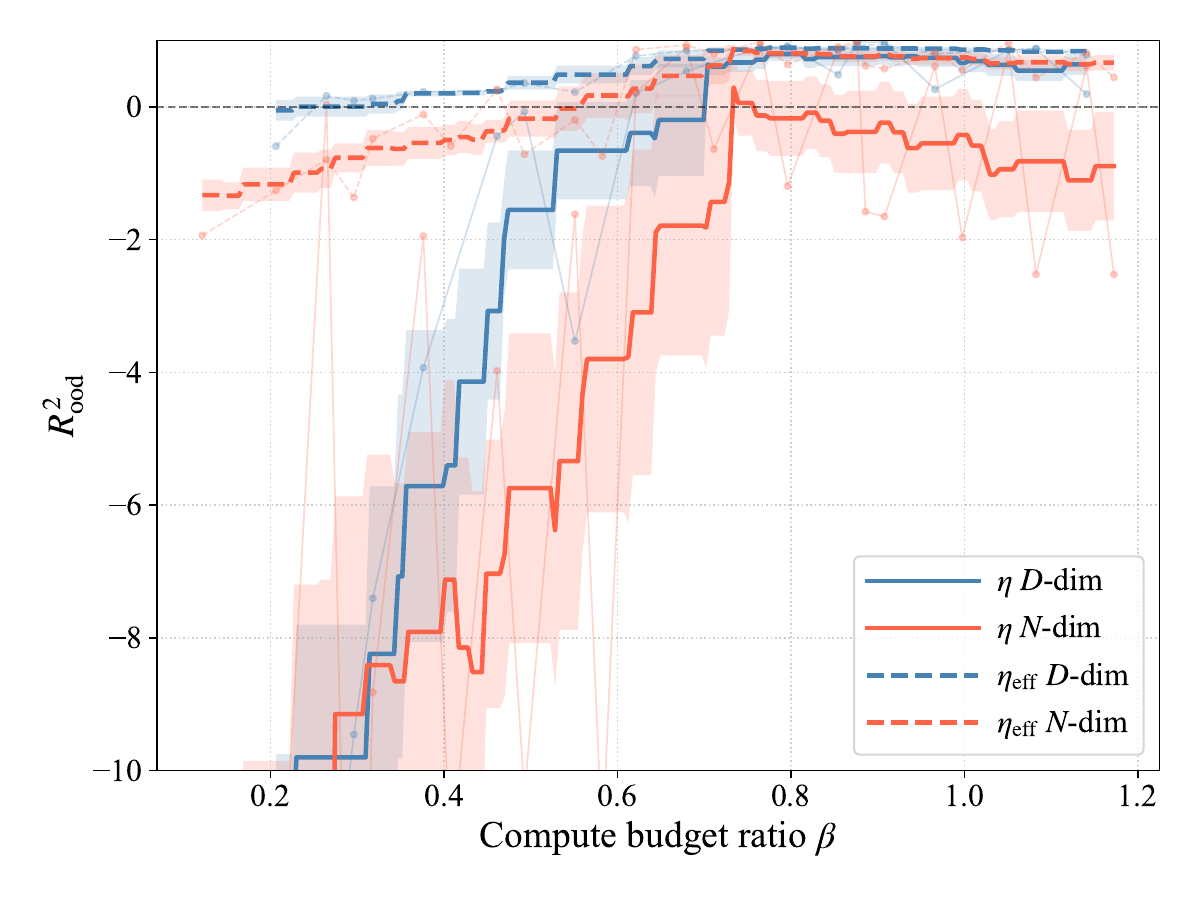}
        \caption{}
        \label{fig:ood}
    \end{subfigure}
    \begin{subfigure}{0.48\textwidth}
        \centering
        \includegraphics[width=\linewidth]{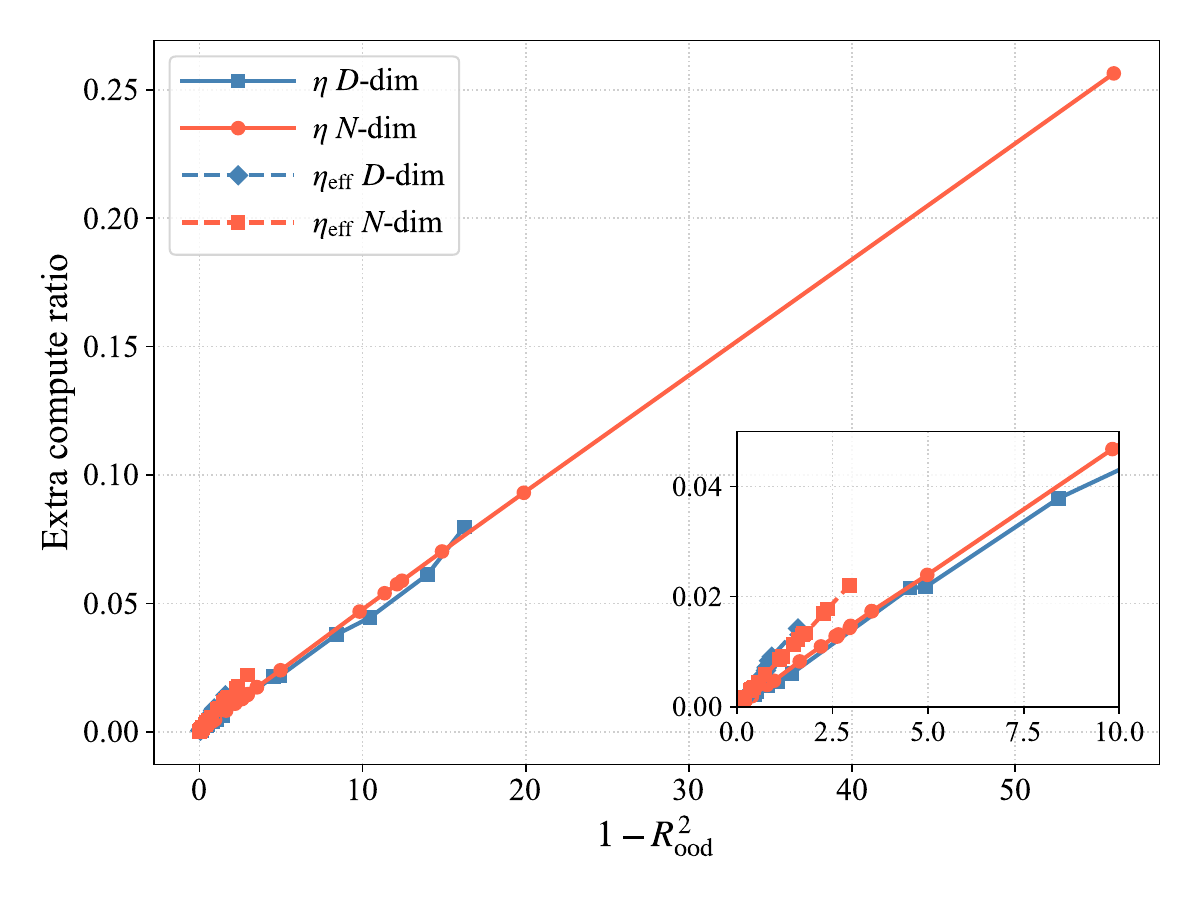}
        \caption{}
        \label{fig:compute_vs_ood}
    \end{subfigure}
    \caption{\textbf{Left}: $R^2_{\text{OOD}}$ across different compute budgets for scaling along $D$ and $N$, using both $\eta^{\ast}$ and $\eta_{\mathrm{eff}}^{\ast}$. Curves are smoothed with a sliding window in linear $\beta$ of half-width $\Delta\beta = 0.1$; bands show $\pm 0.5\sigma$ within each window. Scaling along the $D$ consistently yields better extrapolation than scaling along $N$, and the effective learning rate further improves performance across all settings. \textbf{Right}: Relationship between OOD generalization error ($1 - R^2_{\text{OOD}}$) and extra compute ratio. Each point corresponds to a test configuration. We see a clear linear relationship: $\mathrm{ECR} \propto 1 - R^2_{\text{OOD}}$.}
\end{figure}

\autoref{fig:ood} compares $R^2_{\text{OOD}}$ across different compute budgets for various scaling strategies. When the compute budget ratio increases, the precision of the prediction also increases, which indicates the non-linearity of learning rates versus $\log D$ and $\log N$. 
We observe that scaling with respect to training steps $D$ consistently outperforms scaling with respect to model size $N$. In addition, across all compute budgets and fitting strategies, $\eta_{\mathrm{eff}}^{\ast}$ consistently achieves higher $R^2_{\text{OOD}}$ than $\eta^{\ast}$ for $D$-dim fit and $N$-dim fit respectively. 

In summary, these results show that the most reliable extrapolation is achieved by fitting along the dimension $D$ and using the effective learning rate $\eta_{\mathrm{eff}}^{\ast}$.

\paragraph{Compute Cost of Extrapolation Error}

To quantify the practical consequences of extrapolation error, we relate prediction accuracy to the extra compute ratio $\mathrm{ECR}(\beta)$.

\autoref{fig:compute_vs_ood} plots $\mathrm{ECR}(\beta)$ against $1 - R^2_{\text{OOD}}(\beta)$. 
Empirically, we observe a linear relationship,
\begin{equation}
    \mathrm{ECR} \propto 1 - R^2_{\text{OOD}},
\end{equation}
indicating that extrapolation error translates directly into additional compute cost.
This relationship implies that even modest degradation in extrapolation quality leads to a proportional increase in required training compute, highlighting the practical importance of accurate scaling laws.

To illustrate the practical impact, we present a concrete case study using a single test configuration, $(D_{\text{test}}, N_{\text{test}}) = (100\mathrm{B}, 707\mathrm{M})$. The detailed results in \autoref{tab:prediction_table} show how differences in extrapolation accuracy translate into practical compute overhead and validation-loss gaps.

\begin{table}[t]
\caption{Prediction error, validation loss gap, and extra compute ratio for each fitting setup across some budget ratios, tested on $(D_{\text{test}}, N_{\text{test}}) = (100\mathrm{B}, 707\mathrm{M})$. Bold indicates the lowest ECR within each $\beta$ block. For some small $\beta$ using $D$-dim fit, there is no $N$-dim correspondence. We always set $\alpha_1D_{\text{test}}=5\mathrm{B}, \alpha_2=\beta-\alpha_1$.}
\centering
\small
\begin{tabular}{ccccc}
\toprule
Budget Ratio (\%) & Setup & Log LR Prediction Error & Val Loss Gap ($10^{-4}$) & ECR (\%) \\
\midrule
$13.00$ & $D$-dim, $\eta_{\text{eff}}$ & $+0.1235$ & $4.77$ & $\mathbf{1.21}$ \\
$13.00$ & $D$-dim, $\eta$              & $-0.8144$ & $66.42$ & $16.96$ \\
\midrule
$17.00$ & $D$-dim, $\eta_{\text{eff}}$ & $+0.1148$ & $4.12$ & $\mathbf{1.04}$ \\
$17.00$ & $D$-dim, $\eta$              & $-0.7295$ & $53.21$ & $13.72$ \\
\midrule
$20.64$ & $D$-dim, $\eta_{\text{eff}}$ & $+0.1271$ & $5.06$ & $\mathbf{1.29}$ \\
$20.64$ & $N$-dim, $\eta_{\text{eff}}$ & $-0.1903$ & $11.04$ & $2.21$ \\
$20.64$ & $D$-dim, $\eta$              & $-0.5736$ & $32.82$ & $8.65$ \\
$20.64$ & $N$-dim, $\eta$              & $-0.5461$ & $29.74$ & $7.87$ \\
\midrule
$37.63$ & $D$-dim, $\eta_{\text{eff}}$ & $+0.0935$ & $2.73$ & $\mathbf{0.67}$ \\
$37.63$ & $N$-dim, $\eta_{\text{eff}}$ & $-0.1320$ & $5.35$ & $1.11$ \\
$37.63$ & $D$-dim, $\eta$              & $-0.2061$ & $4.21$ & $1.20$ \\
$37.63$ & $N$-dim, $\eta$              & $-0.2328$ & $5.37$ & $1.52$ \\
\midrule
$55.10$ & $D$-dim, $\eta_{\text{eff}}$ & $+0.0983$ & $3.02$ & $\mathbf{0.75}$ \\
$55.10$ & $N$-dim, $\eta_{\text{eff}}$ & $-0.1357$ & $5.64$ & $1.17$ \\
$55.10$ & $D$-dim, $\eta$              & $-0.2286$ & $5.18$ & $1.47$ \\
$55.10$ & $N$-dim, $\eta$              & $-0.2179$ & $4.71$ & $1.34$ \\
\bottomrule
\end{tabular}
\newline
\label{tab:prediction_table}
\end{table}

\section{Explaining Nonlinear Scaling via Implicit Effective Learning Rate Schedule}
\label{sec:effective_lr_schedule}
In our experiment, we observe that the optimal effective learning rates extrapolates more reliably than standard learning rate when $t$ is large. We explain why optimal learning rate would \textbf{bend upwards} as the training horizon increases. 

Our explanation rests on one assumption: \textit{due to scale-invariance, the effective learning rate $\eta_{\text{eff}}$, instead of the learning rate $\eta$,  governs optimization dynamics}. Form this perspective, $\eta_{\text{eff}}$ is the quantity for which a clean scaling law should hold, and any apparent irregularity in $\eta$ reflects the time-varying mapping between the two. To examine this hypothesis, we first study the case of Adam without weight decay.

\begin{figure}[htbp]
    \centering

    \begin{subfigure}{0.48\textwidth}
        \centering
        \includegraphics[width=\linewidth]{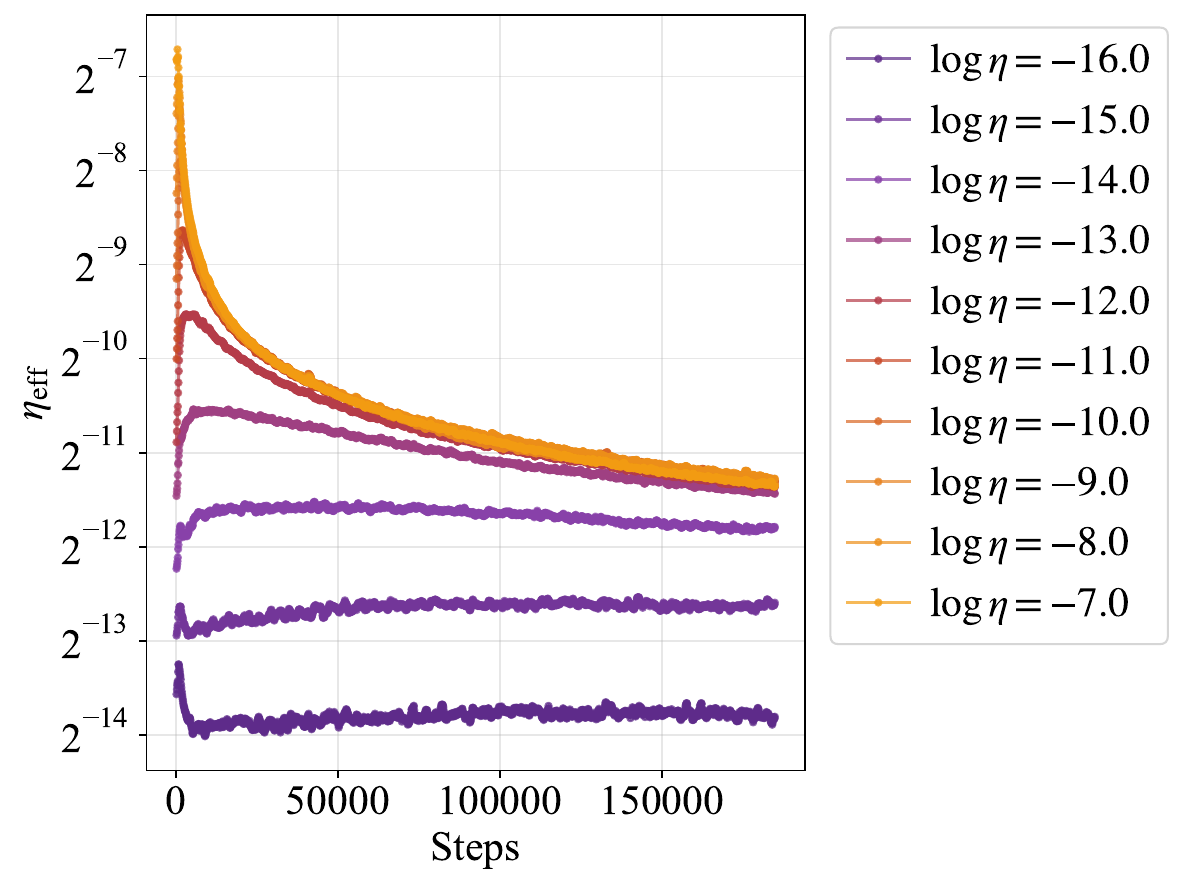}
        \caption{}
        \label{fig:efflr_adam_nowd}
    \end{subfigure}
    \begin{subfigure}{0.48\textwidth}
        \centering
        \includegraphics[width=\linewidth]
        {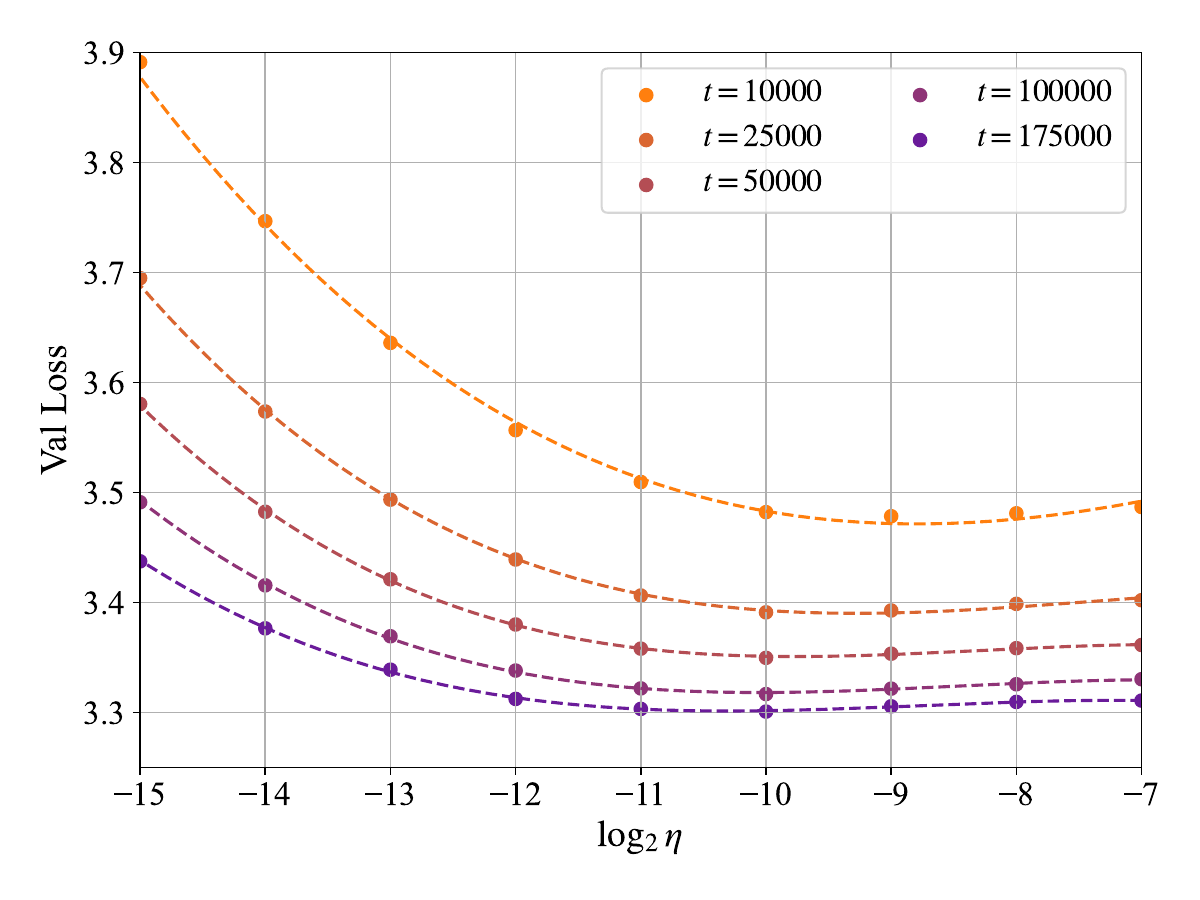}
        \caption{}
        \label{fig:val_loss_adam_nowd}
    \end{subfigure}

    \caption{
        Validation loss and (smoothed) effective learning-rate dynamics for Adam without weight decay. The effective learning rate is computed from the $Q$ matrix in the first attention layer. Experiments use $N = 64\mathrm{M}$ models trained with constant learning rates after a 1000-step warmup.
        \textbf{Left}: For large $\eta$, the trajectories of $\eta_{\text{eff}}(t)$ coincide at large $t$.
        \textbf{Right}: For large $\eta$, validation loss versus $\log_2 \eta$ becomes nearly flat across different training steps.
        }
\end{figure}

\paragraph{Validating The Assumption.} The evolution of $\|w_t\|_2$ with a constant $\eta$ has already been derived from previous work~\citep{kosson2024rotational, wen2025hyperball, kosson2025wdimportant}. \textit{If we train models without applying weight decay}, the effective learning rate decays over time: 
\begin{equation}
    \label{eq:adam_nowd_efflr}
    \eta_{\text{eff}}(t) \approx \frac{1}{\sqrt{\frac{1 + \beta_1}{1 - \beta_1}t + \frac{W_0^2}{\eta^2 U^2}}}
    \sim \sqrt{\frac{1-\beta_1}{1+\beta_1}\frac{1}{t}},
\end{equation}
where the asymptotic form holds for sufficiently large $t$. The key observation is that for sufficiently large $\eta$, the second term in the denominator becomes negligible, so $\eta_{\text{eff}}(t)$, and hence the accumulated step size $\sum_t \eta_{\text{eff}}(t)$, converges to the same trajectory regardless of $\eta$ (as shown in~\autoref{fig:efflr_adam_nowd}). Since training runs with similar accumulated step sizes are known to reach similar losses~\citep{tissue2024scaling, luo2025multipower}, using different large $\eta$ should yield similar validation losses according to our assumption. \autoref{fig:val_loss_adam_nowd} confirms this prediction, which is aloTs consistent with observations from~\citep[Figure~1]{kosson2025wdimportant}; see \autoref{app:experimental_setup} for experimental details.

\paragraph{Two LR-Rate Phases For AdamW.} The equilibrium of weight norms has been shown to play an important role in AdamW\citep{kosson2025wdimportant}. In this work, we want to show that the time it takes for the weight norm to reach equilibrium matters.
Concretely, since $\eta_{\text{eff}}$ depends on the weight norm $\|w_t\|_2$, and $\|w_t\|_2$ requires many steps to relax towards its equilibrium value, small $\eta$ keeps training in a transient regime where the $\eta\leftrightarrow\eta_{\text{eff}}$ relationship differs from its equilibrium form. We now make this precise.

Similar to~\autoref{eq:adam_nowd_efflr}, the evolution of $\|w_t\|_2$ under AdamW with constant $\eta$ is:
\begin{equation}
\label{eq:adamw_efflr}
    \eta_{\text{eff}}(t) = \sqrt{2\eta\lambda\frac{1-\beta_1}{1+\beta_1}\frac{1}{1 + \left(\frac{W_0^2}{W_{+\infty}^2} - 1\right)e^{-2\eta\lambda t}}},
\end{equation}
where $W_0 = \|w_0\|_2$ is the initial weight norm and $W_{+\infty} = U\sqrt{\frac{\eta}{2\lambda}\frac{1 + \beta_1}{1 - \beta_1}}$ is the equilibrium weight norm, with $U$ denoting the (approximately constant~\cite{kosson2024rotational, liu2025muon}) Adam update norm $\|u_t\|_2$. Convergence to equilibrium occurs on a decay time $(2\eta\lambda)^{-1}$, so smaller $\eta\lambda$ produces prolonged transient dynamics in $\eta_{\text{eff}}(t)$. For instance, with $\eta = 2^{-16}$ and $\lambda = 0.1$, the decay time is $327{,}680$ steps, which is comparable to or larger than typical training horizons. The detailed mathematical derivations largely follow prior work \citep{kosson2024rotational, wen2025hyperball, kosson2025wdimportant, li2020reconciling} and are provided in \cref{app:theory} for completeness.

Then we see that \autoref{eq:adamw_efflr} thus exhibits two asymptotic regimes:

\paragraph{Equilibrium regime ($t \gg (2\eta\lambda)^{-1}$).} The exponential term vanishes, giving
\begin{equation}
    \eta_{\text{eff}} \propto \sqrt{\eta},
    \quad \text{i.e.,} \quad
    \log \eta = 2 \log \eta_{\text{eff}} + \mathrm{const}.
\end{equation}

\paragraph{Pre-equilibrium regime ($t \ll (2\eta\lambda)^{-1}$).} The weight-norm term dominates, giving
\begin{equation}
    \eta_{\text{eff}} \propto \eta,
    \quad \text{i.e.,} \quad
    \log \eta = \log \eta_{\text{eff}} + \mathrm{const}.
\end{equation}

\begin{figure}[t]
    \centering
    \includegraphics[width=\linewidth]{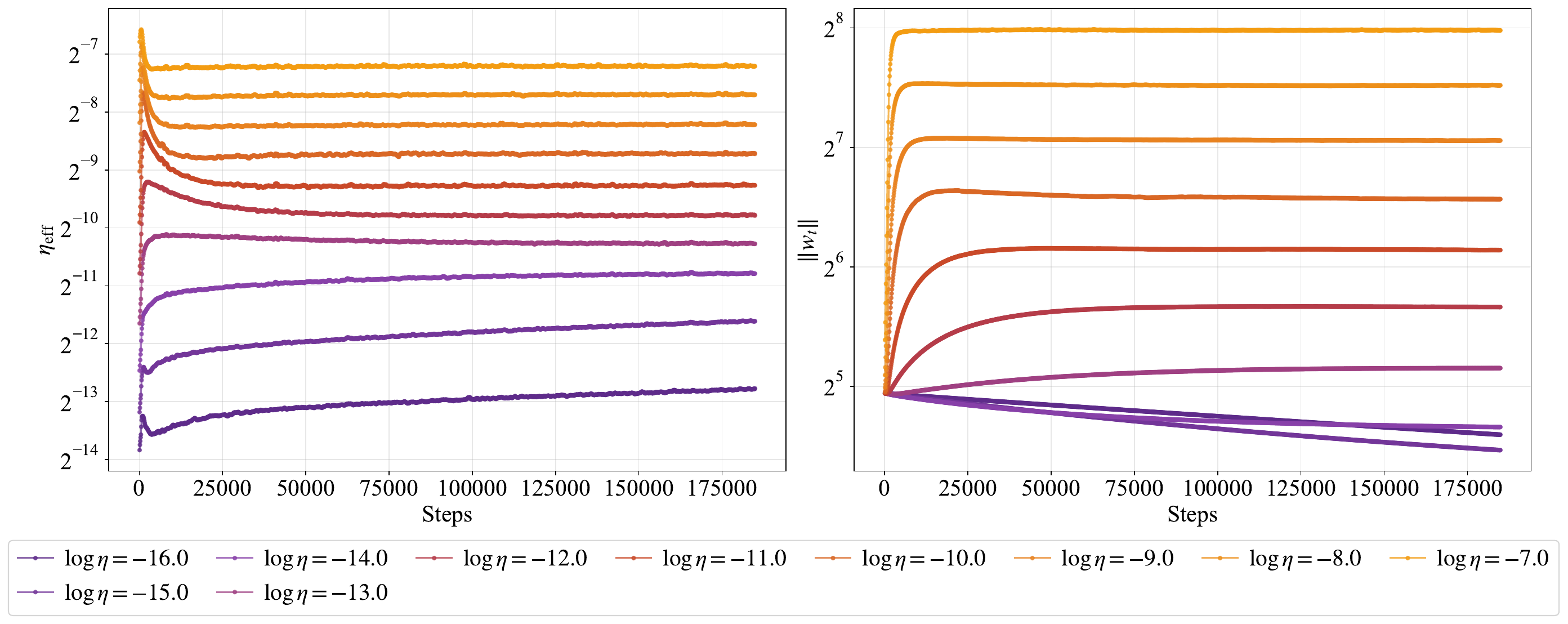}
    \caption{(Smoothed) Evolution of the effective learning rate $\eta_{\mathrm{eff}}(t)$ and weight norm $\|w_t\|_2$, from the $Q$ matrix in the first attention layer. We train $N = 417\mathrm{M}$ with constant learning rates after a 1000-step warmup. The $y$-axis is shown on a log scale.
    \textbf{Left}: $\eta_{\mathrm{eff}}$ exhibits two regimes; adjacent curves are spaced by approximately $0.5$ in the first regime and $1$ in the second.
    \textbf{Right}: The weight-norm trajectory saturates exponentially toward equilibrium, with smaller $\eta$ producing slower convergence.}
    \label{fig:efflr_weight_norm_vs_t}
\end{figure}

Both regimes are clearly visible in \autoref{fig:efflr_weight_norm_vs_t}. The smaller $\eta$ relaxes more slowly toward equilibrium and therefore sits in the pre-equilibrium regime, where $\log\eta$ versus $\log\eta_{\text{eff}}$ has a slope $1$; larger $\eta$ reaches equilibrium quickly and follows the slope-$2$ relation. As the training horizon grows, the optimal learning rate decreases, \textit{pushing training from the equilibrium regime into the pre-equilibrium regime}. This transition explains the upward curvature in \autoref{eq:eta_d_scaling_law}.

\begin{figure}[htbp]
    \centering

    \begin{subfigure}{0.48\textwidth}
        \centering
        \includegraphics[width=\linewidth]{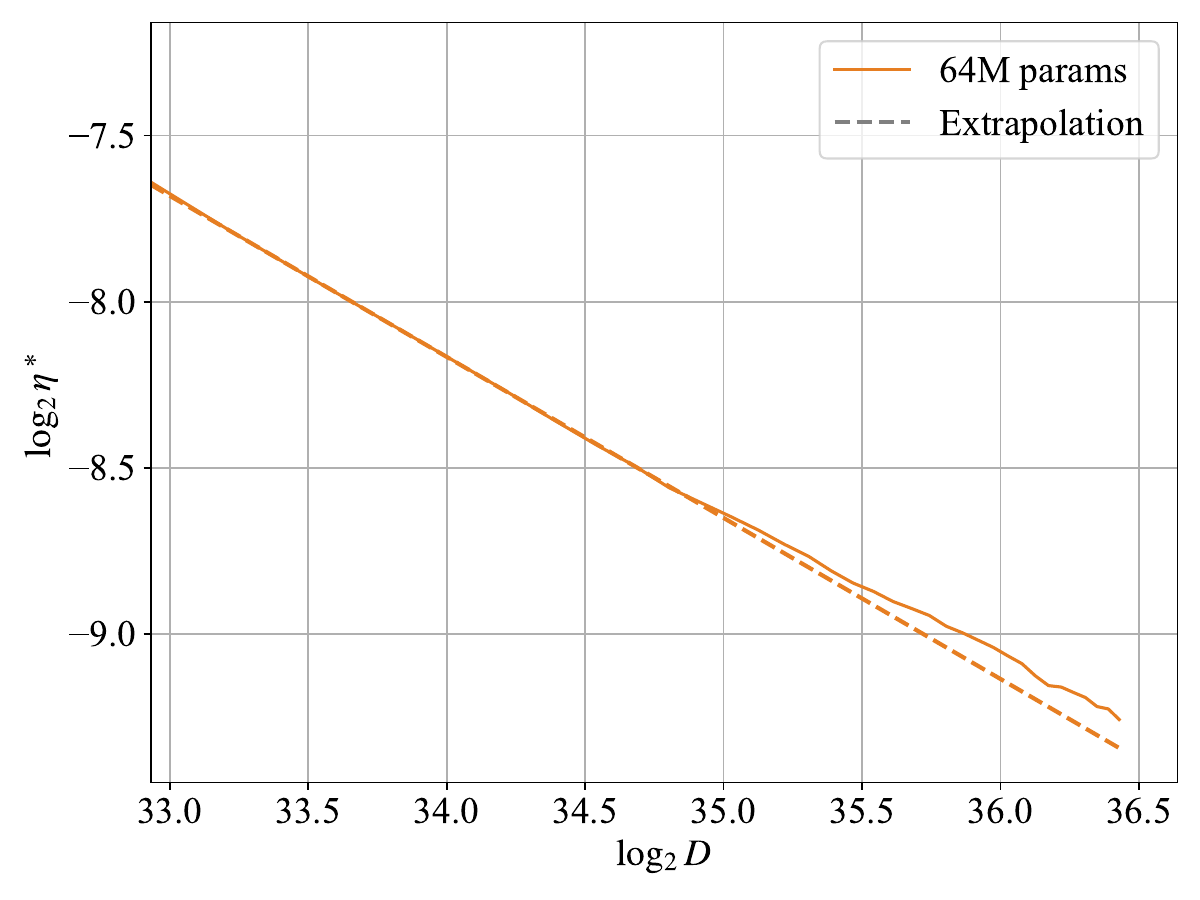}
    \end{subfigure}
    \begin{subfigure}{0.48\textwidth}
        \centering
        \includegraphics[width=\linewidth]{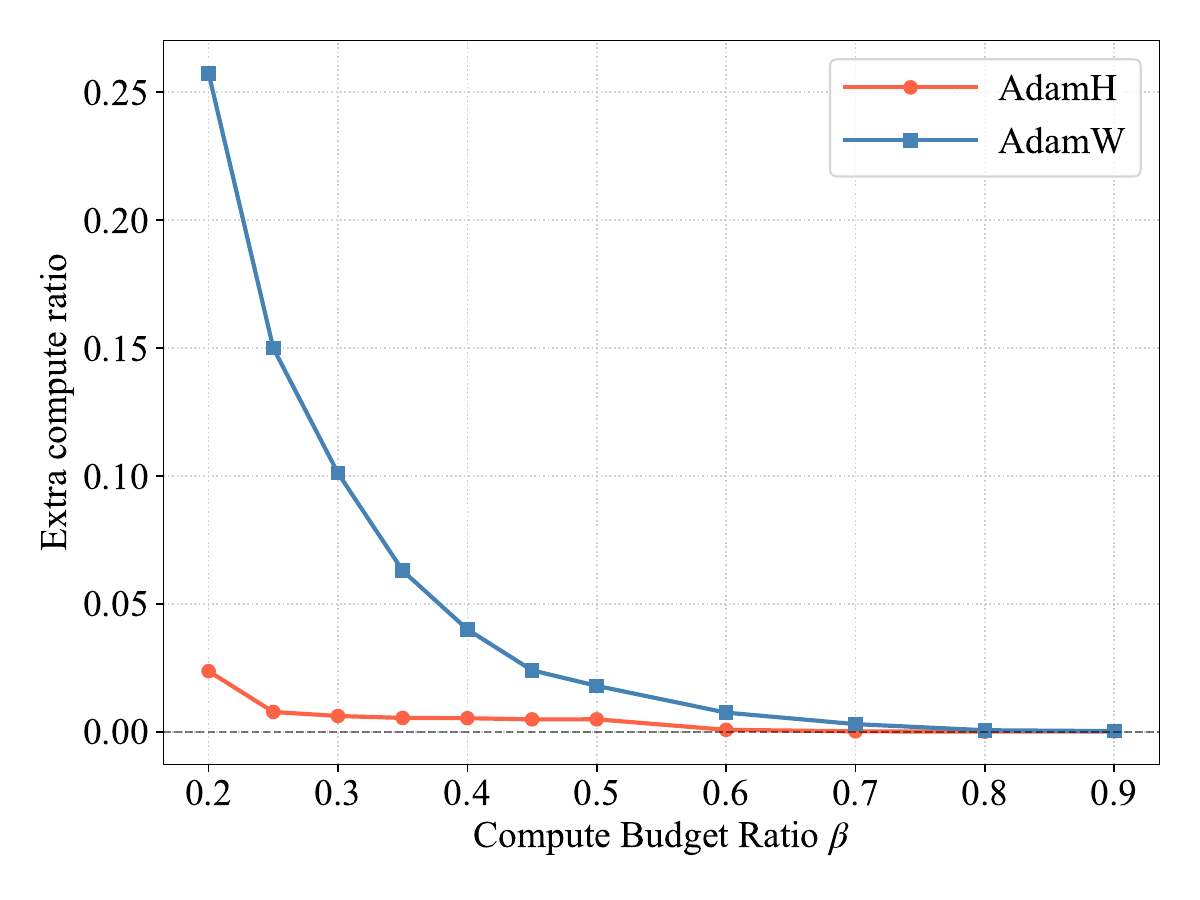}
    \end{subfigure}

    \caption{
    Scaling behavior of AdamH and comparison of ECR with AdamW (Experiments on $N = 64\mathrm{M}$).
    \textbf{Left}: AdamH exhibits a clear scaling relationship, $\log \eta_{\mathrm{eff}}^\ast \propto \log D$. The extrapolation line is fitted by $D \in [20\mathrm{B}, 40\mathrm{B}]$. Pearson coefficent $r=-0.9994$.
    \textbf{Right}: We compare the extra compute ratio (ECR) for $(D_{\text{test}}, N_{\text{test}}) = (100\mathrm{B}, 64\mathrm{M})$. AdamH achieves a much smaller ECR than AdamW at small compute-budget ratios $\beta$, indicating that AdamH follows a more linear scaling law. 
    }
    \label{fig:adamh}
\end{figure}

\paragraph{Explicit Control Effective Learning Rates via AdamH.} Although $\eta_{\text{eff}}$ exhibits a clean power law with $D$, it is not directly controlled during training. Instead, it emerges implicitly from optimizer dynamics. We therefore adopt AdamH~\citep{wen2025hyperball} (\autoref{alg:adamh}), an optimizer that controls $\eta_{\text{eff}}$ explicitly by normalizing both the weight and the update to a fixed norm. Repeating the scaling analysis under AdamH, we recover the same log-linear relationship $\log \eta_{\text{eff}}^* \propto \log D$ (\autoref{fig:adamh}); see \autoref{app:experimental_setup} for details.

\section{Conclusions and Limitations}
\label{sec:limitations}

The contributions of this work are threefold. First, we show that the commonly used log-linear scaling law for the optimal learning rate can exhibit systematic upward curvature at larger scales, leading to underprediction when extrapolating from smaller runs. Second, we identify two choices that substantially improve transfer: parameterization by the effective learning rate, and extrapolating along the data axis rather than the model-size axis. Third, we provide a mechanistic explanation for the observed curvature: under AdamW, the mapping from the raw learning rate to the effective learning rate is shaped by weight-norm dynamics, whose transient behavior becomes more important when the optimal step size is small. Experiments with AdamH, which directly controls the effective learning rate, further support this interpretation.

Several limitations should be acknowledged. \emph{(i) Model and optimizer coverage:} our study uses GPT-2–style architectures trained on FineWeb with AdamW under fixed hyperparameters; generalization to other architectures and optimizers remains to be verified. \emph{(ii) Estimation of $\eta^\ast$:} we rely on cubic fitting to determine $\eta^\ast$, which may introduce bias when the true optimum lies near the sweep boundary.  \emph{(iii) Limited observed scales:} our analysis is limited to scales up to 100B data and 707M model parameters, and does not probe more extreme regimes, where additional phenomena may emerge.

%%%%%%%%%%%%%%%%%%%%%%%%%%%%%%%%%%%%%%%%%%%%%%%%%%%%%%%%%%%%

\newpage 

\bibliography{reference}
\bibliographystyle{icml2026}
%%%%%%%%%%%%%%%%%%%%%%%%%%%%%%%%%%%%%%%%%%%%%%%%%%%%%%%%%%%%
\newpage
\section*{Appendix}

\appendix

\section{Additional Related Works on Hyperparameter Tuning and Transfer}
\label{app:additional_related_work}

Beyond learning rate, recent work has also investigated tuning and transfer strategies for other hyperparameters.
From a theoretical perspective, \citet{wang2025wdscaling} proposes scaling rules for weight decay by formulating model weights as an exponential moving average of optimization updates; subsequent works further demonstrate the role of weight decay in learning-rate transfer \citep{fan2025properwd, kosson2025wdimportant}.
\citet{mccandlish2018empirical, zhang2025cbsscaling} study critical batch size and its scaling behavior.
\citet{jastrzkebski2017sgdsde, Adamlrbsscaling, mareksmall} study how learning rate and momentum should scale with batch size.
More recently, \citet{wang2025sharpness} studies block-wise learning rates in Transformer architectures.

For data-driven approaches, scaling laws have been used to fit and predict optimal hyperparameters beyond learning rate, including batch size \citep{hu2024minicpm, deepseekai2024deepseekscaling, wang2024meituanlaw}. In the context of AdamW~\citep{loshchilov2018decoupled} optimization, they have also been used to tune weight decay \citep{powerline, li2025efficient}.
\citet{tissue2024scaling, luo2025multipower} study how to predict loss curves from learning-rate schedules and identify optimal schedules.
A more classical line of work tackles hyperparameter optimization through Bayesian optimization and related sequential model-based optimization methods \citep{snoek2012practical, hutter2011sequential}.
More recent work further explores learned surrogate performance models \citep{athanasiadis2025tune, zhang2026configuration}.
Synthesizing insights from prior work, \citet{mlodozeniec2025completed} proposes a joint transfer strategy for multiple hyperparameters.

\section{Detailed Training Setups}
\label{app:experimental_setup}

All of our experiments were conducted on 16 H200 GPUs. Including interruptions caused by various intermediate issues, the experiments took approximately two months in total.
\paragraph{Training Details of Main Experiments}
All experiments are conducted using a 12-layer GPT-2 architecture with the attention head dimension fixed at 128. Our training code is based on nanoGPT~\citep{karpathy_nanogpt}, an open-source implementation for training and fine-tuning GPT-style models. You can check our implementation in \url{https://anonymous.4open.science/r/LRTransfer}. We vary the scale of the model by changing the hidden dimension over $\{256,512,768,1024,1536,2048\}$, resulting in models with parameter counts ranging from 22M to 707M. All models are initialized following the scheme described in the original GPT-2 paper~\citep{radford2019gpt2}.

By default, weight matrices, including the token embedding matrix and linear projection matrices, are initialized from a normal distribution $\mathcal{N}(0, 0.02)$.
As an exception, projection matrices in residual branches are initialized with a scaled
standard deviation $0.02 / \sqrt{2L}$, where $L$ is the number of Transformer layers.

Models are trained on FineWeb-100B with batch size 512, sequence length 1024 and vocabulary size 50304, corresponding to approximately 0.52M tokens per optimization step. We train for up to 200k steps, giving a maximum data budget of approximately 100B tokens. Optimization uses AdamW with weight decay $0.1$, momentum parameters $\beta_1=\beta_2=0.95$, and $\epsilon=10^{-8}$. For the main model parameters, we sweep the learning rate over a logarithmic grid defined by $\log_2 \eta \in [-15,-7]$ with unit increments.

For the embedding layer, we use a separate AdamW optimizer with fixed learning rate $0.0036$, weight decay $0.1$, $\beta_1=0.9$, $\beta_2=0.95$, and $\epsilon=10^{-8}$.

Training follows a warmup--steady--decay (WSD) schedule. We use a 1000-step warmup followed by a steady training phase. To evaluate performance at different data budgets, we periodically save checkpoints throughout the steady phase, allowing us to reconstruct intermediate training states without re-running experiments. 

Using these checkpoints, we derive alternative training configurations with a longer decay phase of 5000 steps, which is used for all results reported in the main experiments. After the decay phase, we evaluate each checkpoint on 10M tokens and compute the validation loss. This procedure allows us to efficiently evaluate multiple training horizons ranging from 10k to 200k steps, corresponding to effective data sizes from 5B to 100B tokens. We organized these training logs into a dataset, \href{https://huggingface.co/datasets/neurips2026lrtransfer/LR-Transfer-Trajectory}{LR-Transfer-Trajectory dataset}.

\paragraph{Training Details for AdamH and Adam Experiments.}
All experimental settings are identical to those used in the main experiments,
including model architecture, dataset (FineWeb-100B), and optimization hyperparameters.
The embedding layer is optimized separately using Adam without weight decay.
For AdamH, the algorithm is illustrated as \autoref{alg:adamh}. The range of the learning rate sweep is $\eta \in [-11, -7]$.
A 5000-step decay has been applied to the results given in \autoref{fig:adamh}.

\begin{algorithm}[H]
\caption{AdamH Optimizer}
\label{alg:adamh}
\begin{algorithmic}[1]
\Require Initial weights $W_0$, learning rate $\eta$, Adam hyperparameters $(\beta_1, \beta_2, \epsilon)$
\State Initialize $m_0 \gets 0$, $v_0 \gets 0$, $R \gets \|W_0\|_2$
\For{$t = 1, 2, \dots, T$}
    \State Compute gradient $g_t \gets \nabla_W \mathcal{L}(W_t)$
    \State $m_t \gets \beta_1 m_{t-1} + (1 - \beta_1) g_t$
    \State $v_t \gets \beta_2 v_{t-1} + (1 - \beta_2) g_t^2$
    \State $\hat{m}_t \gets m_t / (1 - \beta_1^t)$
    \State $\hat{v}_t \gets v_t / (1 - \beta_2^t)$
    \State $u_t \gets \hat{m}_t / (\sqrt{\hat{v}_t} + \epsilon)$ \Comment{Adam update}
    \State $\tilde{W}_{t+1} \gets W_t - \eta R \cdot \mathrm{Normalize}(u_t)$
    \State $W_{t+1} \gets R \cdot \mathrm{Normalize}(\tilde{W}_{t+1})$
\EndFor
\end{algorithmic}
\end{algorithm}

\section{The implicit Schedule of The Effective Learning Rate}
\label{app:theory}

Similar analysis can be find in~\citep{kosson2024rotational,kosson2025wdimportant,wen2025hyperball}. Here we just derive these formulas in our notation.

The AdamW update is
\begin{equation}
    w_{t+1} = w_t - \eta (u_t + \lambda w_t) = \alpha w_t - \eta u_t, \quad \alpha = 1 - \eta\lambda.
\end{equation}
Let $W_t = \|w_t\|_2$ denote the weight norm.

\paragraph{Assumptions.}
We make two simplifying assumptions about the dynamics, as in \citep{wen2025hyperball}:
\begin{itemize}
    \item The update norm stabilizes quickly during training: $\|u_t\|_2 \approx U$.
    \item Temporal correlations decay geometrically: $\langle u_t, u_{t'} \rangle \approx U^2 \beta_1^{|t-t'|}$.
\end{itemize}
The first is an empirical observation~\citep{liu2025muon}, while the second follows from the momentum structure. 

\paragraph{Weight Norm Dynamics.}
Unrolling the recursion expresses $w_t$ as a weighted sum of past updates:
\begin{equation}
    w_t = -\eta \sum_{i=0}^{t-1} \alpha^{t-1-i} u_i.
\end{equation}
Taking inner product with $u_t$ and using the correlation assumption, we obtain
\begin{equation}
    \label{eq:correlation}
    \langle u_t, w_t \rangle \approx -\eta U^2 \sum_{i=0}^{t-1} \alpha^{t-1-i}\beta_1^{t-i}
    \approx -\frac{\eta\beta_1}{1-\alpha\beta_1}U^2,
\end{equation}
where the second step follows from summing a geometric series.

We now compute how the norm evolves. Expanding the squared norm of the update,
\begin{equation}
    W_{t+1}^2 = \|\alpha w_t - \eta u_t\|_2^2 = \alpha^2 W_t^2 + \eta^2 U^2 - 2\alpha\eta \langle w_t,u_t\rangle.
\end{equation}
Substituting the expression above yields a closed recursion:
\begin{equation}
    W_{t+1}^2 \approx \alpha^2 W_t^2 + \eta^2 \frac{1+\alpha\beta_1}{1-\alpha\beta_1}U^2.
\end{equation}

This recursion can be solved explicitly. When $\lambda>0$,
\begin{equation}
    W_t^2 = \frac{\eta^2}{1-\alpha^2}\frac{1+\alpha\beta_1}{1-\alpha\beta_1}U^2(1-\alpha^{2t}) + W_0^2\alpha^{2t}.
\end{equation}
For $\eta\lambda \ll 1$, we use $\alpha^{2t} \approx e^{-2\eta\lambda t}$ to obtain
\begin{equation}
    \label{eq:wt_with_wd}
    W_t^2 \approx \frac{\eta}{2\lambda}\frac{1+\beta_1}{1-\beta_1}U^2(1-e^{-2\eta\lambda t}) + W_0^2 e^{-2\eta\lambda t}.
\end{equation}
This shows that the weight norm converges exponentially to
\begin{equation}
    W_{+\infty} = U\sqrt{\frac{\eta}{2\lambda}\frac{1+\beta_1}{1-\beta_1}},
\end{equation}
with a characteristic decay time $(2\eta\lambda)^{-1}$.

When $\lambda=0$, the recursion simplifies and yields linear growth:
\begin{equation}
    W_t^2 = \eta^2\frac{1+\beta_1}{1-\beta_1}U^2 t + W_0^2.
\end{equation}

\paragraph{Effective Learning Rate Evolution.}
According to the definition of the effective learning rate~\autoref{eq:efflr_definition}:
\begin{equation}
    \eta_{\text{eff}}(t) = \|\hat{w}_{t+1}-\hat{w}_t\|_2, \quad \hat{w}_t = \frac{w_t}{\|w_t\|}.
\end{equation}
Using the cosine formula,
\begin{equation}
    \eta_{\text{eff}}(t) = \sqrt{2 - 2\frac{\langle w_t,w_{t+1}\rangle}{W_t W_{t+1}}},
\end{equation}
and substituting
\begin{equation}
    \langle w_t,w_{t+1}\rangle = \alpha W_t^2 - \eta \langle w_t,u_t\rangle,
\end{equation}
we obtain
\begin{equation}
    \eta_{\text{eff}}(t) = \sqrt{2\left(1 - \alpha\frac{W_t}{W_{t+1}} - \frac{\eta^2\beta_1}{1-\alpha\beta_1}\frac{U^2}{W_t W_{t+1}}\right)}.
\end{equation}

For sufficiently large $t$, we have $W_t\approx W_{t+1}$. This expression simplifies to
\begin{equation}
    \eta_{\text{eff}}(t) \approx \frac{\eta U}{W_t},
\end{equation}
showing that the effective learning rate is inversely proportional to the weight norm.

Substituting the solutions for $W_t$, we obtain closed-form expressions.

When $\lambda>0$,
\begin{equation}
    \eta_{\text{eff}}(t) \approx \sqrt{2\eta\lambda\frac{1-\beta_1}{1+\beta_1}\frac{1}{1 + \left(\frac{W_0^2}{W_{+\infty}^2} - 1\right)e^{-2\eta\lambda t}}}.
\end{equation}

When $\lambda=0$,
\begin{equation}
    \eta_{\text{eff}}(t) \approx \frac{1}{\sqrt{\frac{1+\beta_1}{1-\beta_1}t + \frac{W_0^2}{\eta^2U^2}}}
    \sim \sqrt{\frac{1-\beta_1}{1+\beta_1}\frac{1}{t}}.
\end{equation}

\paragraph{Validating The Theory}
We validate the theory with data collected from the $Q$ matrix of the first attention layer, with the model scale $N=124\mathrm{M}$.

\begin{figure}[htbp]
    \centering
    \includegraphics[width=0.75\linewidth]{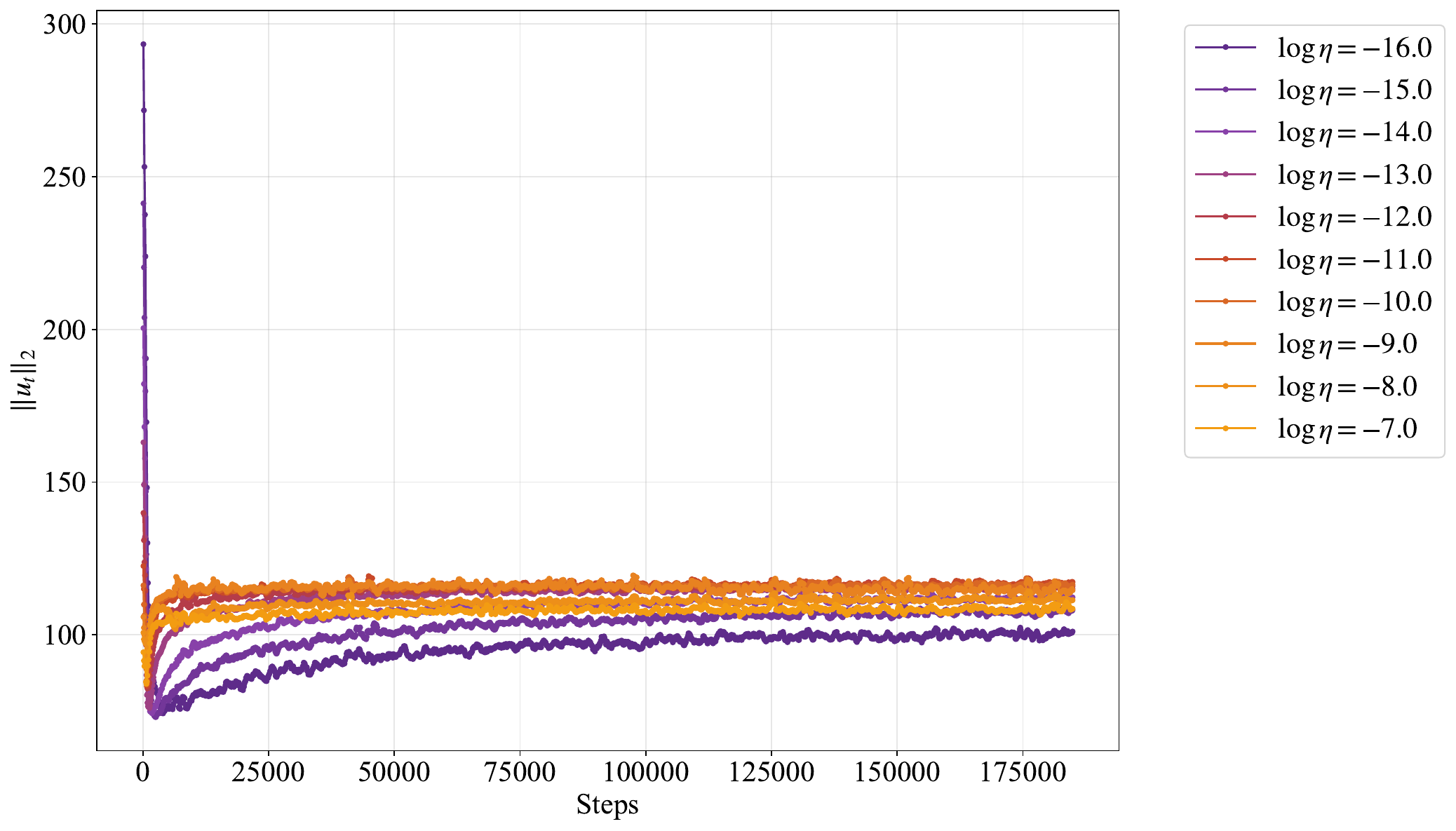}
    \caption{(Smoothed) Evolution of the $\|u_t\|_2$.}
    \label{fig:optimizer_norm_vs_t}
\end{figure}

First, let us test the first assumption. As illustrated in \autoref{fig:optimizer_norm_vs_t}, this assumption holds well for large learning rates. For small learning rates, it still holds approximately, since $\|u_t\|_2$ varies slowly.

\begin{figure}[htbp]
    \centering
    \includegraphics[width=0.75\linewidth]{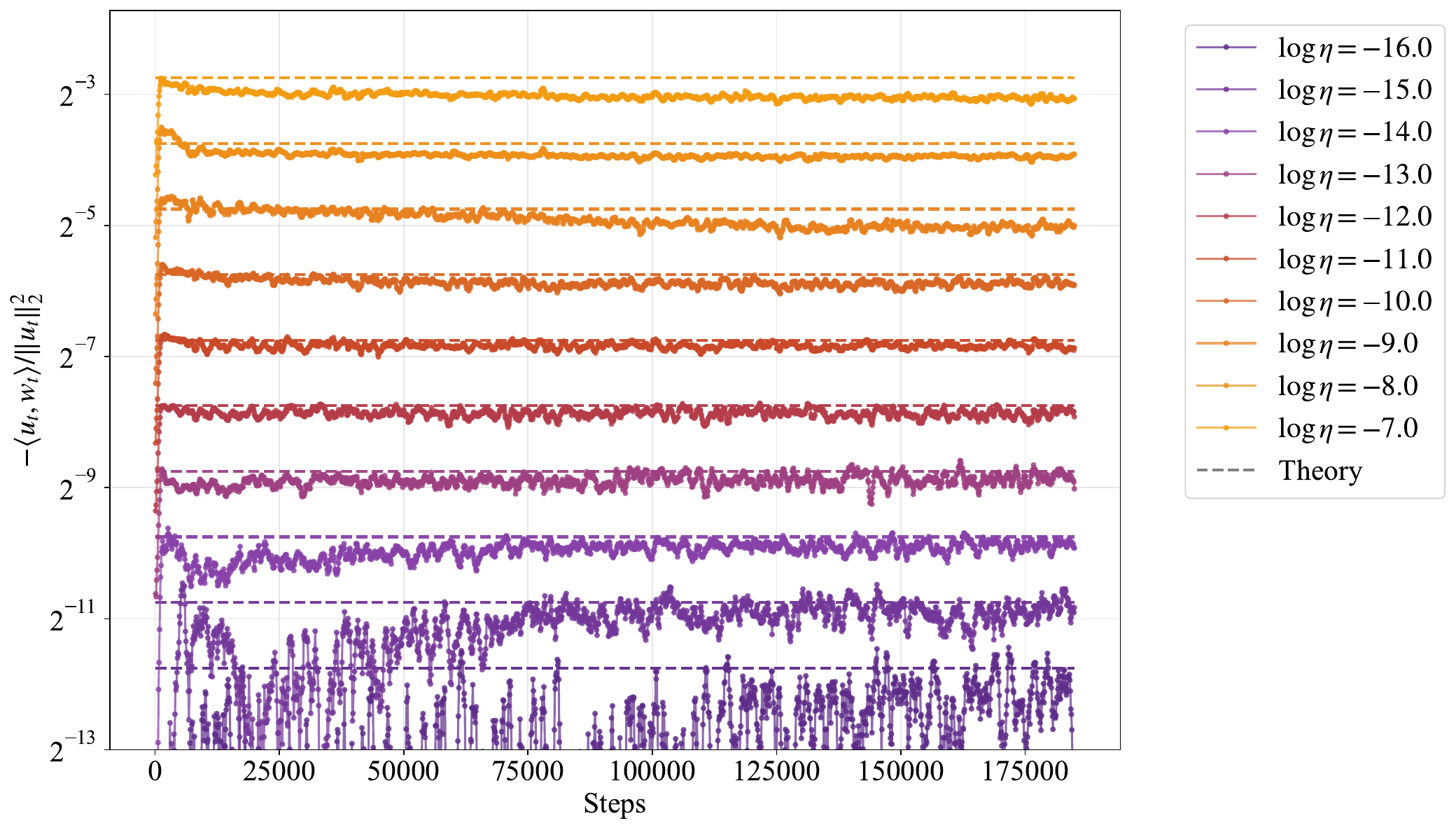}
    \caption{(Smoothed) Evolution of the $-\frac{\langle u_t, w_t\rangle}{\|u_t\|_2}$.}
    \label{fig:correlation_vs_t}
\end{figure}

Next, for the second assumption, it's hard to test it directly. Hence, we examine it by justifying \autoref{eq:correlation}. From \autoref{fig:correlation_vs_t}, we find that the experimental results are very close to the prediction of the theory.

\begin{figure}[htbp]
    \centering
    \includegraphics[width=0.75\linewidth]{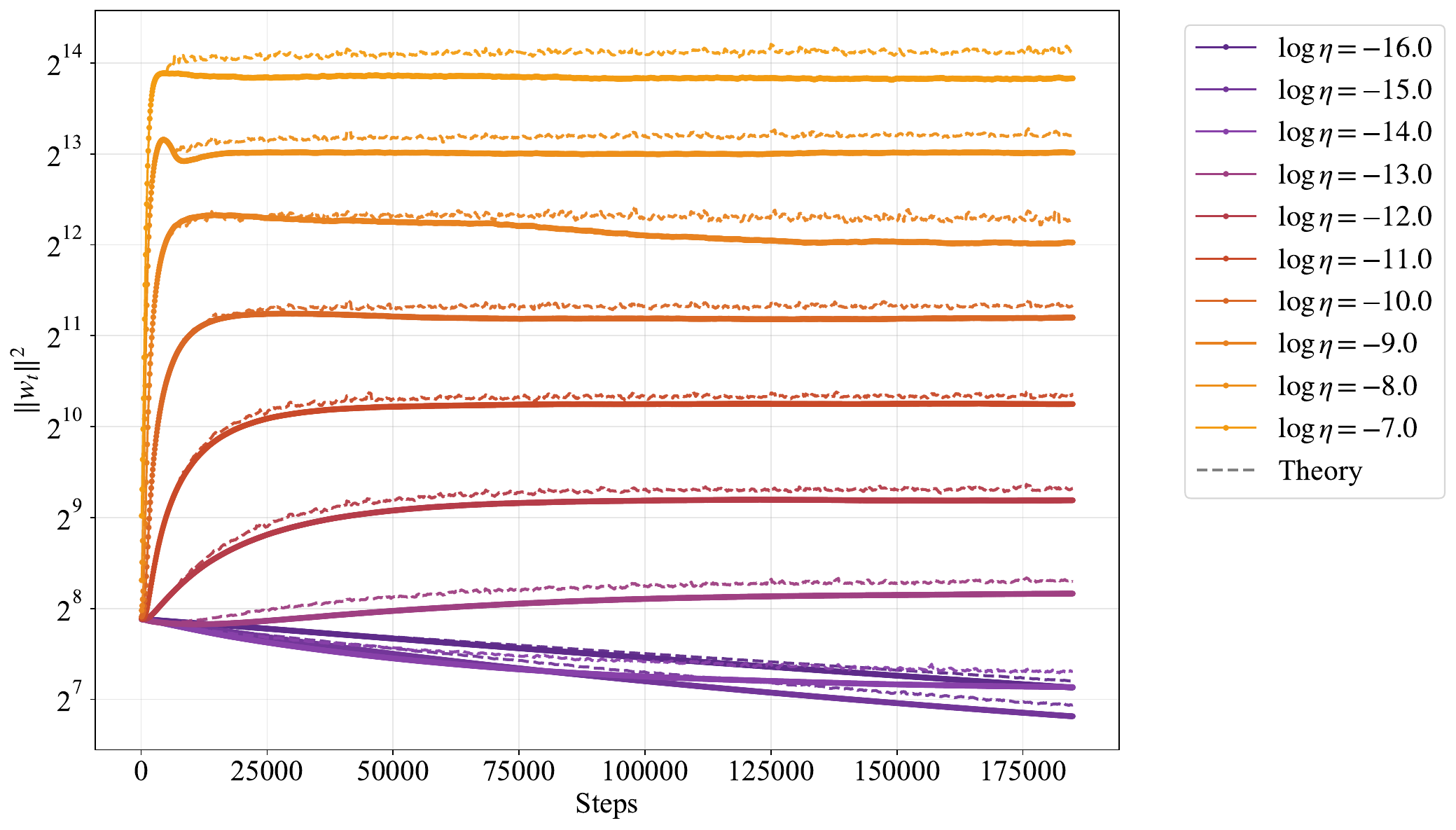}
    \caption{(Smoothed) Evolution of the $\|w_t\|^2$.}
    \label{fig:wt_square_vs_t}
\end{figure}

Finally, we test how $\|w_t\|_2^2$ changes over time. As $\|u_t\|_2$ changes slowly, we can use $\|u_t\|_2^2$ to replace $U^2$ in \autoref{eq:wt_with_wd}. The result of the prediction is shown in \autoref{fig:wt_square_vs_t}.

\newpage

\section{More Results}

\paragraph{Cubic Fit Results}

\begin{figure}[h]
    \centering
    \includegraphics[width=0.95\linewidth]{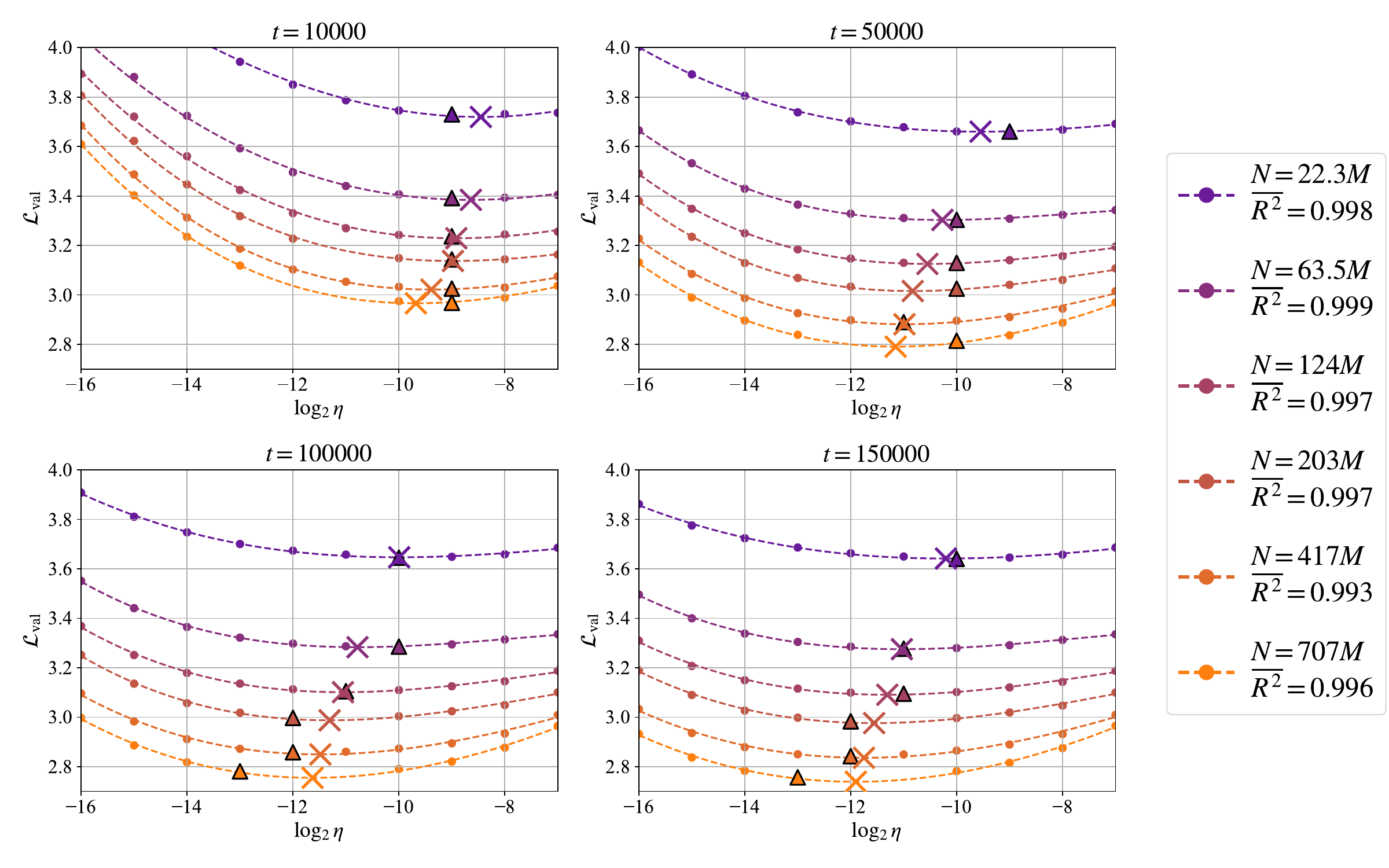}
    \caption{Validation loss fitted as a cubic function of $\log \eta$ for different steps.}
    \label{fig:val_loss_vs_log2lr}
\end{figure}

\begin{figure}[h]
    \centering
    \includegraphics[width=0.95\linewidth]{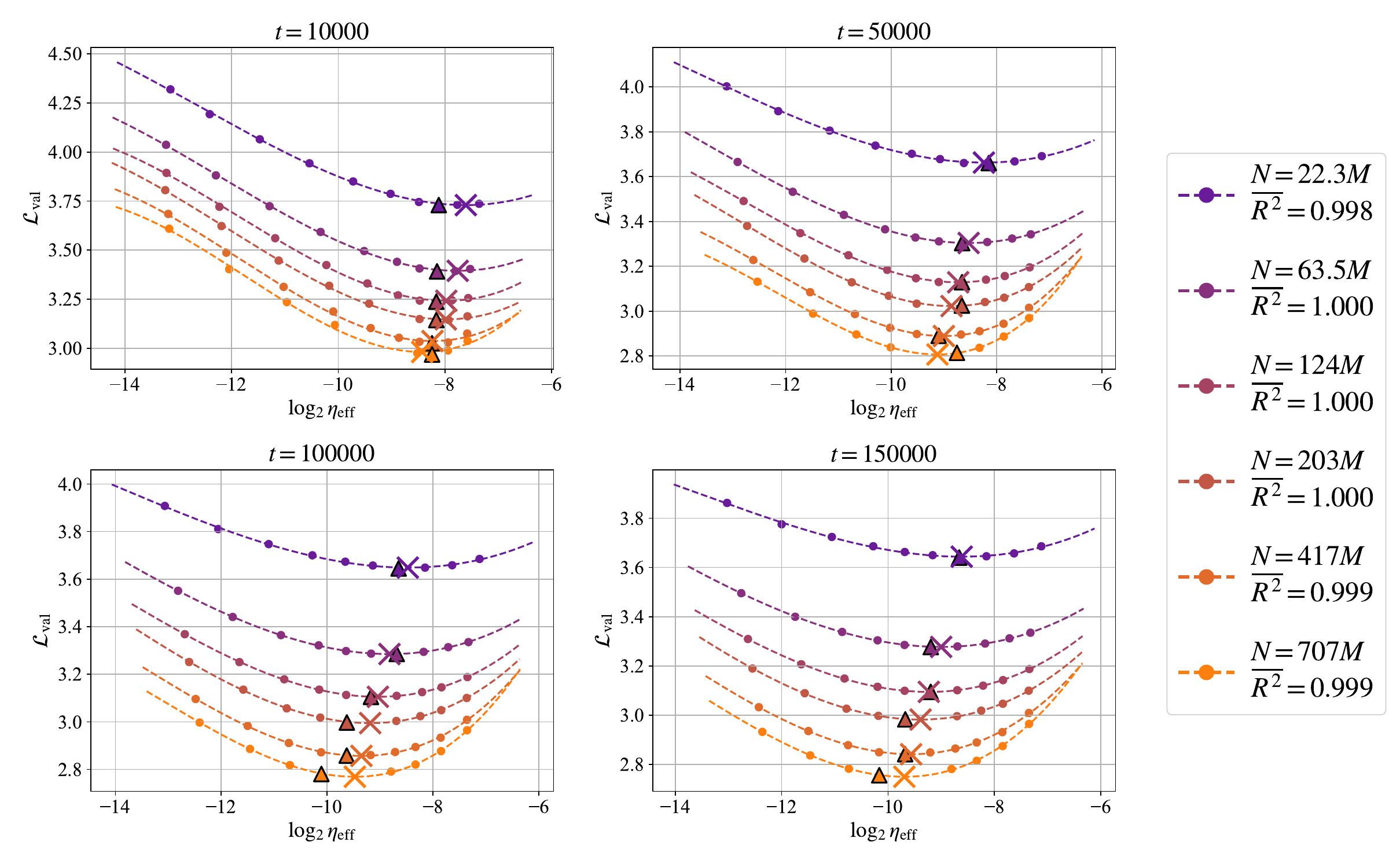}
    \caption{Validation loss fitted as a quadratic function of $\log \eta_{\text{eff}}$ for different steps.}
    \label{fig:val_loss_vs_log2efflr}
\end{figure}

For \autoref{fig:val_loss_vs_log2lr} and \autoref{fig:val_loss_vs_log2efflr}, we use circles to denote the raw data points from the learning rate sweep, and the dashed curve represents the cubic fit. $R^2 = 1 - \frac{\sum_{i=1}^{n} (y_i - \hat{y}_i)^2}
{\sum_{i=1}^{n} (y_i - \bar{y})^2}$ indicates goodness of fit. The triangle marks the lowest validation loss observed among the raw data, while the $\times$ indicates the minimizer of the fitted curve, which we take as the optimal learning rates. We observe a clear asymmetry in the validation loss on the two sides of the extremum: it increases more rapidly for smaller learning rates. This justifies our use of a cubic fit.

\begin{figure}[h]
    \centering
    \includegraphics[width=0.95\linewidth]{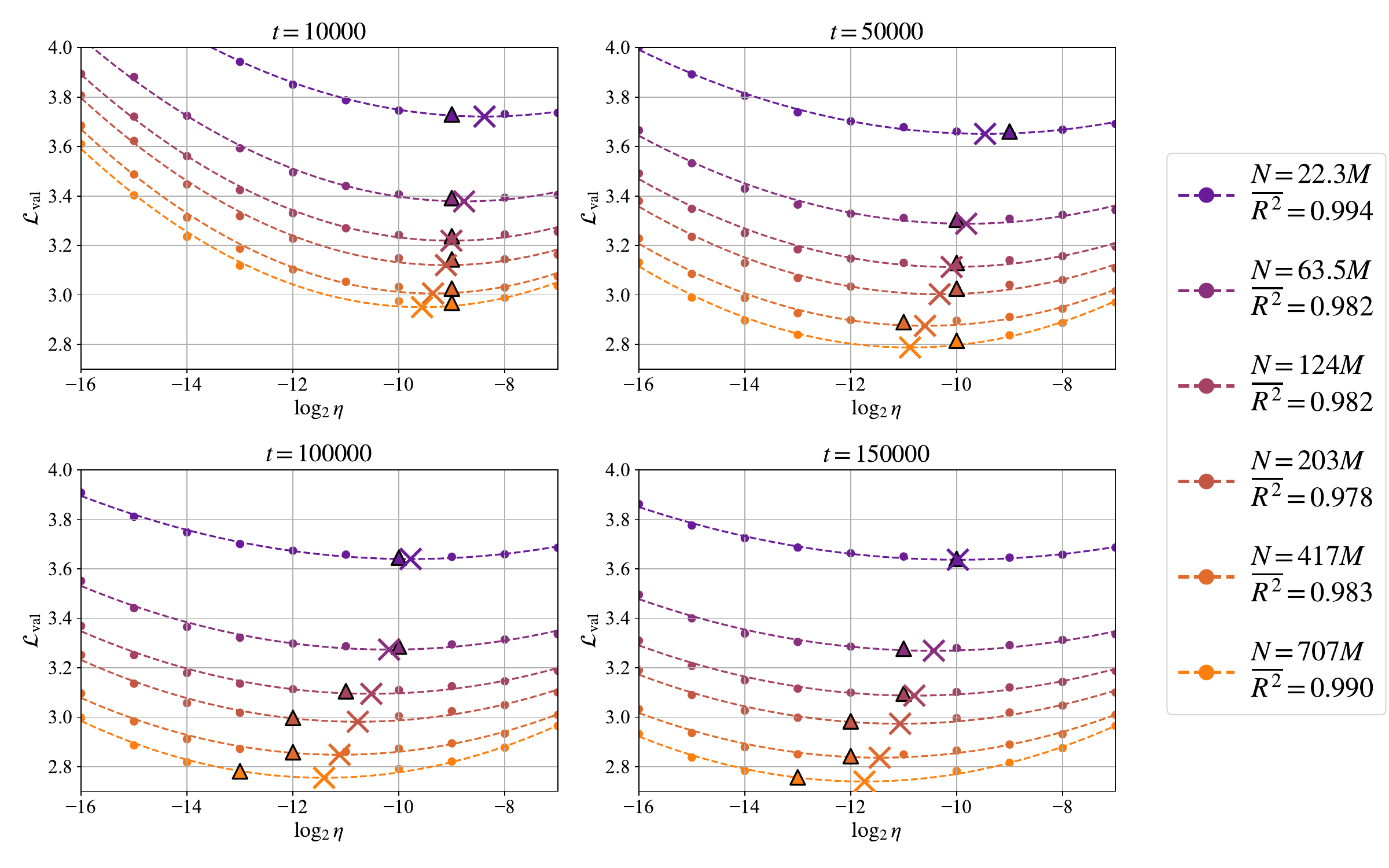}
    \caption{Validation loss fitted as a quadratic function of $\log \eta$ for different steps.}
    \label{fig:val_loss_vs_log2lr_quadratic}
\end{figure}

\begin{figure}[h]
    \centering
    \includegraphics[width=0.95\linewidth]{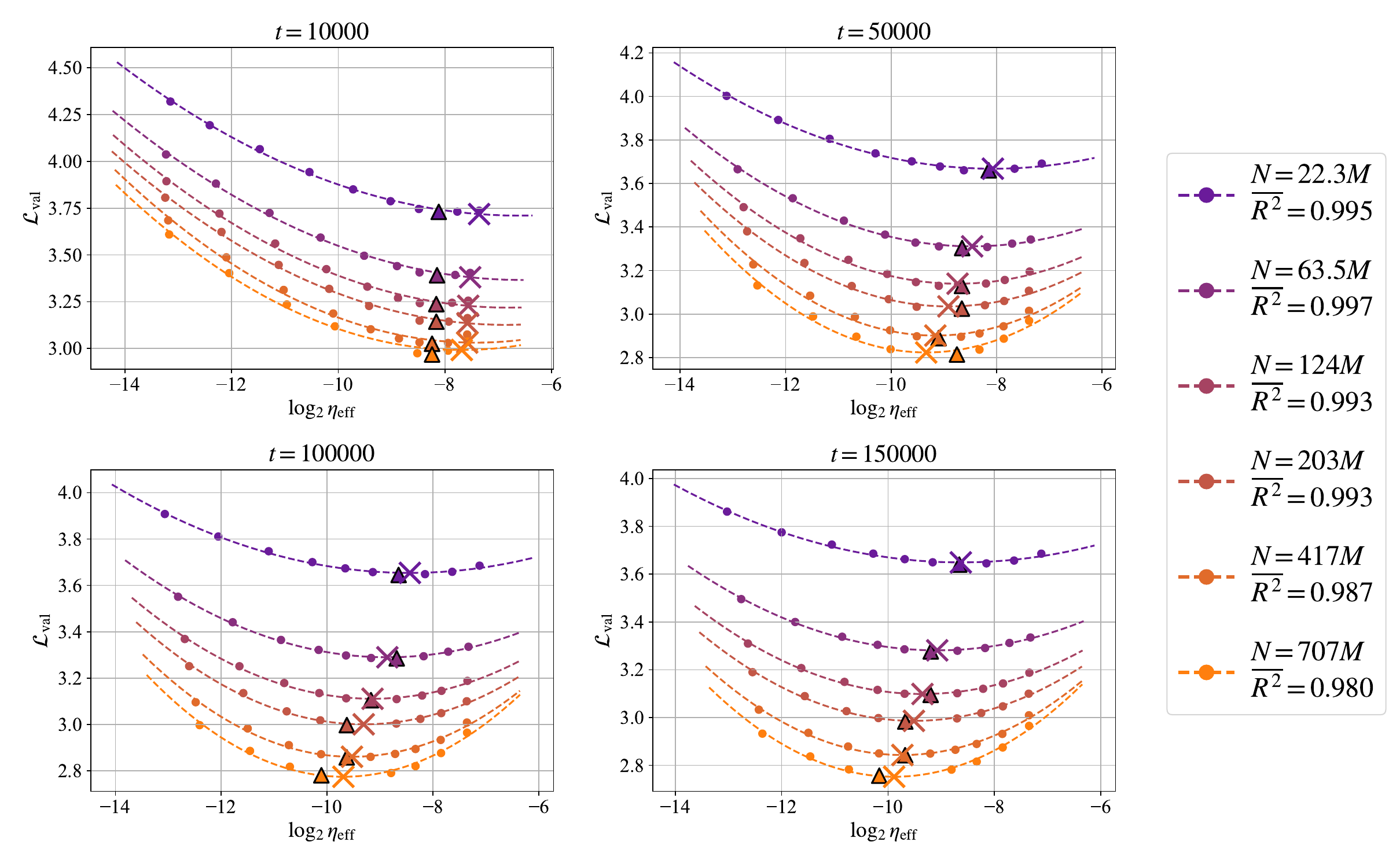}
    \caption{Validation loss fitted as a quadratic function of $\log \eta_{\text{eff}}$ for different steps.}
    \label{fig:val_loss_vs_log2efflr_quadratic}
\end{figure}

\paragraph{Quadratic Fit Results} In comparison, we show the performance of a quadratic fit in \autoref{fig:val_loss_vs_log2lr_quadratic} and \autoref{fig:val_loss_vs_log2efflr_quadratic}. We see that the $R²$ of the quadratic fit is lower than that of the cubic fit.

\newpage

\paragraph{Weight-wise Differences in Effective Learning Rates}

\begin{figure}[h]
    \centering
    \includegraphics[width=\linewidth]{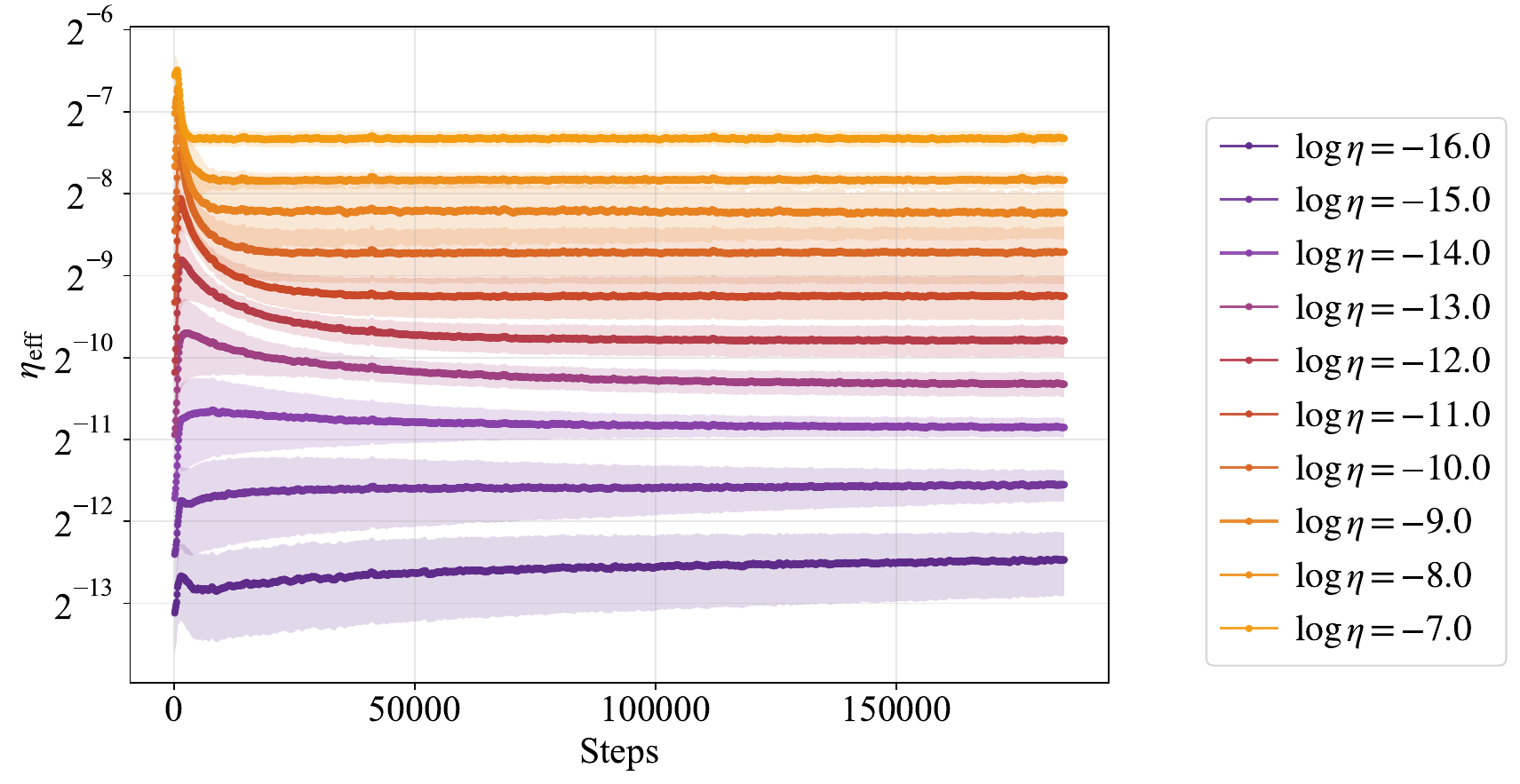}
    \caption{Effective learning rates averaged over all weights (mean $\pm$ half a standard deviation) for $N=124\mathrm{M}$.}
    \label{fig:efflr_difference}
\end{figure}

In the main text, all effective learning rates are averaged over different weights. From ~\autoref{fig:efflr_difference}, we see that there is a weight difference in effective learning rates. After taking the average, the two regimes still exist.

\end{document}